%% file: main.tex
\documentclass[sigconf,9pt]{acmart}
\usepackage[utf8]{inputenc}

\usepackage{amsmath,amssymb,amsfonts}
\usepackage{algorithmic}
\usepackage{graphicx}
\usepackage{textcomp}
\usepackage{xcolor}
\usepackage{booktabs} 
\usepackage{soul}
\usepackage{subcaption}
\usepackage[bb=boondox]{mathalfa}
\usepackage{dsfont} 
\usepackage{extarrows} 
\usepackage{hyperref}
\usepackage{stmaryrd}
\usepackage{gensymb}
\usepackage{afterpage}
\usepackage{enumitem}
\usepackage{natbib}
\usepackage{graphicx}
\usepackage[ruled,linesnumbered]{algorithm2e}

\newlength\mylenin
\newcommand\myinput[1]{%
\settowidth\mylenin{\KwIn{}}%
\setlength\hangindent{\mylenin}%
\hspace*{\mylenin}#1\\}

\let\oldnl\nl
\newcommand{\nonl}{\renewcommand{\nl}{\let\nl\oldnl}}

\newlength\mylenout

\newcommand{\blue}[1]{\textcolor{black}{#1}} 

\newcommand{\brown}[1]{\textcolor{black}{#1}} 

\newcommand{\green}[1]{\textcolor{black}{#1}} 

\newcommand{\tool}{\texttt{HAPI}}

\newif\ifcomment

\commenttrue
\commentfalse

\ifcomment
\definecolor{stelios_colour}{RGB}{144, 238, 144}
\newcommand{\stelios}[1]{\sethlcolor{stelios_colour}\hl{[\textbf{Stelios:} #1]}}
\newcommand{\nic}[1]{\sethlcolor{yellow}\hl{[\textbf{Nic:} #1]}}
\newcommand{\hj}[1]{\sethlcolor{brown}\hl{[\textbf{Hyeji:} #1]}}
\newcommand{\steve}[1]{\sethlcolor{cyan}\hl{[\textbf{Stefanos:} #1]}}
\else
\newcommand{\stelios}[1]{}
\newcommand{\steve}[1]{}
\newcommand{\hj}[1]{}
\newcommand{\nic}[1]{}
\fi

\makeatletter
\def\footnoterule{\relax%
  \kern-5pt
  \hbox to \columnwidth{\hfill\vrule width 1\columnwidth height 0.4pt\hfill}
  \kern4.6pt}
\makeatother

\AtBeginDocument{%
  \providecommand\BibTeX{{%
    \normalfont B\kern-0.5em{\scshape i\kern-0.25em b}\kern-0.8em\TeX}}}
    
\setcopyright{acmcopyright}
\copyrightyear{2020}
\acmYear{2020}
\acmConference[ICCAD '20]{The 39th International Conference on Computer-Aided Design}{November 2--5, 2020}{San Diego, USA}
\acmBooktitle{The 39th International Conference on Computer-Aided Design (ICCAD '20), November 2--5, 2020, San Diego, USA}
\acmPrice{15.00}
\acmDOI{10.1145/XXXXXXX.XXXXXXX}
\acmISBN{XXXXX}
    
\setcopyright{none}

\usepackage{background}
\backgroundsetup{angle=0, 
scale=1,
color=black,
firstpage=true,
position=current page.north,
hshift=0pt,
vshift=-20pt,
contents={\ifnum\value{page}=1 PREPRINT: Accepted at the 39th International Conference on Computer-Aided Design (ICCAD), 2020 \else \fi}
}

\title{HAPI: Hardware-Aware Progressive Inference}

\author{
{Stefanos Laskaridis$^\dagger$*, Stylianos I. Venieris$^\dagger$*, Hyeji Kim$^\dagger$, Nicholas D. Lane$^{\dagger,\ddagger}$}}

\authorwithoutinstuition{{Stefanos Laskaridis, Stylianos I. Venieris, Hyeji Kim, Nicholas D. Lane}}
	
\affiliation{\institution{$^\dagger$Samsung AI Center, Cambridge\hspace{+0.75cm}$^\ddagger$University of Cambridge}{\Small\textit{{* Indicates equal contribution.}}}}

\input{abstract.tex}

\begin{document}

\copyrightyear{2020}
\acmYear{2020}
\setcopyright{acmcopyright}\acmConference[ICCAD '20]{IEEE/ACM International Conference on Computer-Aided Design}{November 2--5, 2020}{Virtual Event, USA}
\acmBooktitle{IEEE/ACM International Conference on Computer-Aided Design (ICCAD '20), November 2--5, 2020, Virtual Event, USA}
\acmPrice{15.00}
\acmDOI{10.1145/3400302.3415698}
\acmISBN{978-1-4503-8026-3/20/11}

\fancyhead{}
\maketitle

\section{Introduction}

\input{intro.tex}

\vspace{-0.1cm}
\section{Background and Related work}
\label{sec:related}
\input{related_work.tex}
\vspace{-0.1cm}
\section{HAPI Overview}
\label{sec:prog_infer}

\input{progressive_inference.tex}

\section{Modelling Framework}
\label{sec:sdf}

\input{sdf_modelling.tex}

\vspace{-0.1cm}
\section{Design Space Exploration}
\label{sec:dse}

\input{design_space_exploration.tex}

\vspace{-0.1cm}
\section{Evaluation}
\label{sec:evaluation}

\input{evaluation.tex}

\vspace{-0.1cm}
\section{Conclusion}

This paper presents a framework for generating optimised progressive inference networks on heterogeneous hardware. By parametrising early-exit networks in a highly customisable manner, the proposed system tailors the number and placement of early exits together with the exit policy to the user-specified performance requirements and target platform. 
Evaluation shows that \tool{} consistently outperforms all baselines \brown{by a significant margin}, demonstrating that i) the design choices are critical in the resulting performance and ii) \tool{} effectively explores the design space and yields a high-performing early-exit network for the target platform.
\vspace{-0.2cm}

\balance
\bibliographystyle{ACM-Reference-Format}
\bibliography{references}
\end{document}

%% file: abstract.tex
\begin{abstract}
Convolutional neural networks (CNNs) have recently become the state-of-the-art in a diversity of AI tasks. 
Despite their popularity, CNN inference still comes at a high computational cost. 
A growing body of work aims to alleviate this by exploiting the difference in the classification difficulty among samples and early-exiting at different stages of the network.
Nevertheless, existing studies on early exiting have primarily focused on the training scheme, without considering the use-case requirements or the deployment platform. 
This work presents \tool{}, a novel methodology for generating high-performance early-exit networks by co-optimising the placement of intermediate exits together with the early-exit strategy at inference time.
Furthermore, we propose an efficient design space exploration algorithm which enables the faster traversal of a large number of alternative architectures and generates the highest-performing \mbox{design}, tailored to the use-case requirements and target hardware.
Quantitative evaluation shows that our system {consistently} outperforms {alternative search mechanisms} and state-of-the-art early-exit schemes across various latency budgets. Moreover, it pushes further the performance of highly optimised hand-crafted early-exit CNNs, delivering up to 5.11$\times$ speedup over lightweight models on imposed latency-driven SLAs for \mbox{embedded devices}. 

\end{abstract}

%% file: intro.tex


Recently, convolutional neural networks (CNNs) have become quintessential for modern intelligent systems; from mobile applications to autonomous robots, CNNs drive 
critical tasks including perception~\cite{maskrcnn2018tpami} and decision making~\cite{kouris2018iros}. 
With an increasing number of CNNs deployed on user-facing setups~\cite{facebook2019}, latency optimisation emerges as a primary objective that can enable the end system to provide low response time. This is also of utmost significance for robotic platforms, to guarantee timely navigation decisions and improve safety, and smartphones to provide smooth user experience. 
%
Nevertheless, despite their unparalleled predictive power, CNNs are also characterised by high inference time due to heavy computational demands, especially when deployed on embedded devices~\cite{embench_2019}.  
To this end, several methods have been proposed to reduce the 
complexity of CNNs and attain minimal latency~\cite{eff_proc_dnns_2017}.

\begin{figure}[t]
    \centering
    \includegraphics[width=1\columnwidth,trim={0.15cm 7.5cm 0.75cm 1cm},clip]{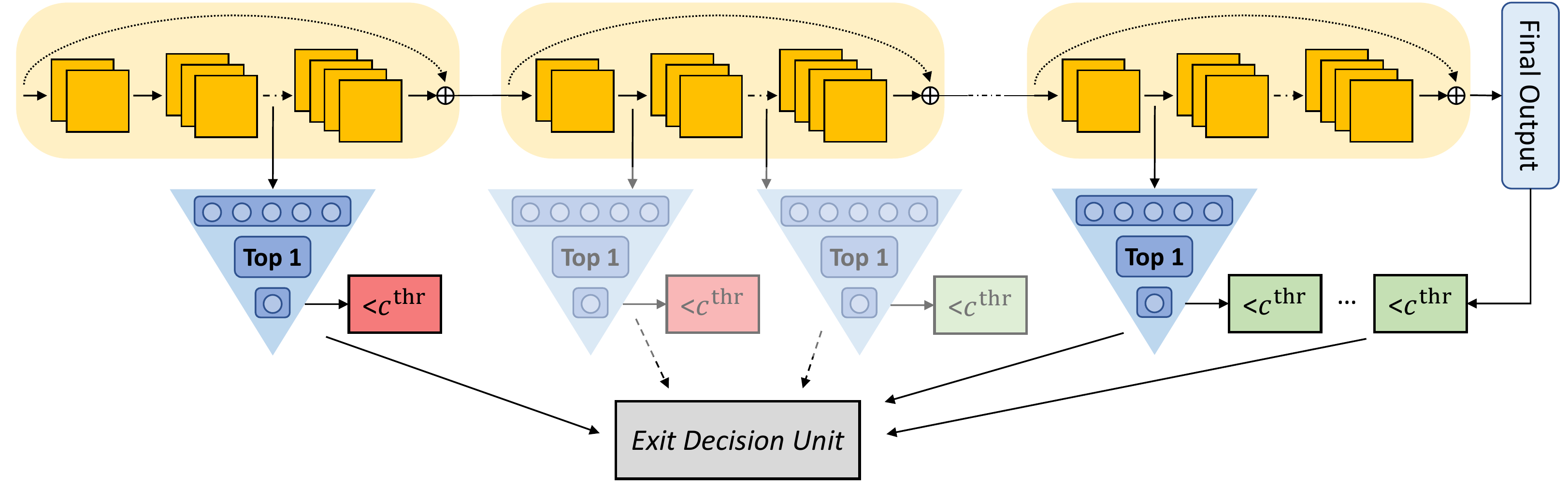}
    \vspace{-0.8cm}
    \caption{\tool{}'s early-exit network deployment architecture.}
    \vspace{-0.6cm}
    \label{InfEngine_v3.pdf}
\end{figure}

Among the existing latency-oriented methods, one line of work focuses on the observation that not all inputs demonstrate the same classification difficulty, and hence each
sample requires different amount of computation to obtain an accurate result.
The conventional techniques that exploit this property typically involve classifier cascades 
\cite{mcdnn_2016,focus_2018,cascadecnn2018,cascadecnn2020date}. 
Despite their effectiveness under certain scenarios, these approaches come with the substantial overhead of deploying and maintaining multiple models. 
%
%
An alternative input-aware approach 
is grounded in the design of early-exit networks~\cite{branchynet2016, Huang2017,scan2019neurips,sdn_icml_2019,spinn2020mobicom}. As illustrated in Fig.~\ref{InfEngine_v3.pdf}, early-exiting takes advantage of the fact that easy samples can be accurately classified using the low-level features that can be found in the earlier layers of a CNN. 
In this manner, each sample would ideally exit the network at the appropriate depth, saving 
time and computational resources.

So far, early-exit works have followed a hardware- and application-agnostic approach, focusing either on the hand-tuned design of early-exit CNN architectures \cite{Huang2017,scan2019neurips} or on optimising the corresponding training scheme \cite{branchynet2016,Zhang_2019_ICCV,sdn_icml_2019}. 
Nonetheless, with CNN-based applications demonstrating diversity both in terms of performance requirements and target processing platforms, tailoring the early-exit network to the use-case needs has remained unexplored. Furthermore, due to the dynamic input-dependent execution and the large space of design choices, the tuning of the early-exit architecture poses a significant challenge that until now required prohibitively long development cycles. \steve{I feel we need some citation here.} 

In this paper, we propose \textbf{\tool{}}, an automated framework that generates an optimised early-exit CNN tailored to the application demands and the target hardware capabilities. 
To generate a high-performance design, \tool{} employs a novel accuracy- and hardware-aware design space exploration (DSE) methodology that enables the efficient traversal over a wide range of \brown{candidate} designs and the effective customisation of the network to the given application-platform pair. 
%
\noindent
The key contributions of this paper are the following:
\vspace{-0.3cm}
\begin{itemize}
    %
    %
    \item {A \emph{Synchronous Dataflow} (SDF) model for representing early-exit CNN workloads and their unique input-dependent dynamic execution. 
    \blue{Our SDF model represents early-exit variants in a dual graph-matrix form that allows us to express the hardware-aware design of an early-exit CNN as a mathematical 
    optimisation problem.}
    More importantly, it enables the previously unattainable fast traversal of the design space by means of algebraic operations that \brown{explore} the accuracy-performance trade-off of the underlying early-exit network implementation.}

    \item {The \tool{} framework for generating progressive inference networks customised for the target deployment platform. The developed framework takes as input a given CNN in PyTorch, performs fast design space exploration by manipulating the SDF model and yields an early-exit implementation customised to meet the user-specified latency target at the maximum accuracy. Through a multi-objective search algorithm, \tool{} explores early-exit designs at both the architectural and exit-policy levels, enabling the rapid adaptation of the target CNN across heterogeneous hardware without the need for retraining, by means of a \textit{train-once, deploy-anywhere} workflow.} 

\end{itemize}

%% file: related_work.tex

Several methods have been proposed for reducing the computational footprint of CNNs in order to speed up computation or fit the model into an embedded device. 
Diverse techniques such as \textit{pruning} \cite{lee2019snip}, \textit{quantisation} \cite{wang2019haq} and \textit{knowledge distillation} \cite{Zagoruyko2017} all aim to reduce the size and latency of a model. 
{
Moreover, NetAdapt~\cite{netadapt2018eccv} also introduces hardware-awareness in its CNN pruning method. However, when a new platform is targeted, the pruned model needs to be fine-tuned through additional high-overhead training iterations. \tool{} employs a single training round upfront, with the per-platform customisation taking place efficiently without training in the loop.}
All these methods are orthogonal to our approach and can be combined together to enable even lower inference cost.

Closer to our approach, cascade systems
also exploit the difference in classification difficulty among inputs to accelerate CNN inference. A cascade of classifiers is typically organised in a multi-stage architecture. Depending on the prediction confidence, the input either exits at the current stage or passes to the next one. 
In this context, several optimisations have been proposed including domain-specific tuning~\cite{focus_2018}, run-time model selection~\mbox{\cite{mcdnn_2016,adapt_model_sel2018lctes,Lee_2019}} and assigning different precision per stage~\cite{cascadecnn2018,cascadecnn2020date}. 
Although these techniques can be effective, the training and maintenance of multiple models add significant overhead to their deployment. In essence, multiple models have to be stored, with a scheduler implementing the model selection logic at inference time. Every time a different model is selected, the system pays the overhead of loading it. 

In contrast to multi-model cascades, a few works have focused on introducing intermediate outputs to a single network. 
BranchyNet \cite{branchynet2016} is an network design with early exits ``branching'' out of the \textit{backbone} architecture of the original network, aiming to speed up inference. While the technique is applicable to various backbone architectures, it was only evaluated on small models and datasets. Moreover, BranchyNet lacks an automated method for tuning the early-exit policy and setting the number and position of exits. 
 
Shallow-Deep Network (SDN)~\cite{sdn_icml_2019} is a more recent work that emphasises the negative impact of 
always exiting at the last exit on accuracy -- a term coined as ``overthinking.'' 
SDN attaches early exits throughout the network and explores the joint training of the exits together with the backbone architecture.
However, the placement of early exits is always equidistant and their number \blue{is fixed to six}, without \blue{optimising for} the task at hand or the device capabilities. Moreover, the degrading effect of early-exit placement to the accuracy of subsequent ones in joint training is not discussed.
Last, although the approach is evaluated on various networks, they do not show any scalability potential to the full ImageNet dataset.


On the other hand, MSDNet~\cite{Huang2017} builds on top of the DenseNet architecture, with each layer working on multiple scales. At each layer, the network maintains multiple filter sizes of diminishing spatial dimensions, but growing depth. These characteristics make the network more robust to placing intermediate classifiers. However, this is a very computationally heavy network, which in turn makes it difficult to deploy on
resource-constrained, latency-critical setups. Moreover, the placement of exits and their co-optimisation during training can hurt the performance of subsequent classifiers, or even lead to instability and non-convergence. Albeit \blue{this challenge has motivated \tool{}'s approach of} decoupling the training of the early exits from the backbone network (see \textit{Early-exit-only training} in Sec.~\ref{sec:training}), subsequent work from the same authors presents techniques for alleviating the limitations of early-exit networks training. Their proposed methodology remains orthogonal to our work~\cite{Li2019a}.

{
We also note that in the existing early-exit approaches, the exit policy and the number and location of the early exits are determined \textit{manually}. \tool{} automates this process by tailoring the early-exit network to the performance requirements and target platform. 
}

%% file: progressive_inference.tex



\begin{figure}[t]
    \centering
    \vspace{-0.25cm}
    \includegraphics[width=1\columnwidth,trim={3cm 3.5cm 3.5cm 1.45cm},clip]{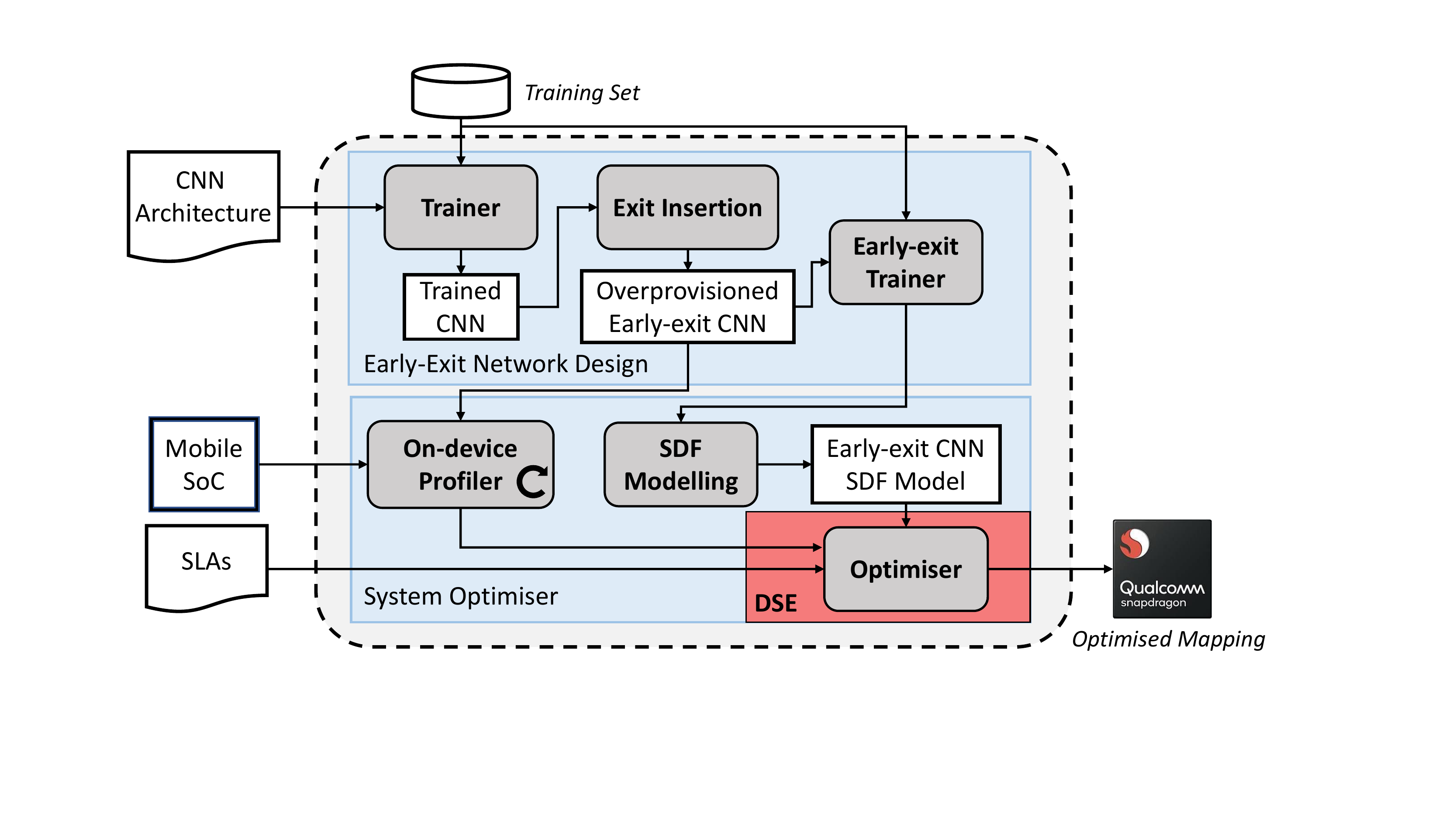}
    \vspace{-0.7cm}
    \caption{Overview of \tool{}'s processing flow.}
    \vspace{-0.5cm}
    \label{HAPI_Arch}
\end{figure}

Fig.~\ref{HAPI_Arch} shows an overview of \tool{}'s processing flow. The framework is supplied with a high-level description of a network
, the task-specific dataset, the target hardware platform and the 
requirements in terms of accuracy, latency and memory. First, if the supplied CNN is not pre-trained, the \textit{Trainer} component trains the network on the supplied training set. 
\blue{Next, the architecture is augmented by inserting intermediate classifiers at all candidate early-exit points, leading to an \textit{overprovisioned} network.}
At this stage, \tool{} freezes the main branch of the CNN and performs {early-exit-only} training (Sec.~\ref{sec:training}).
\green{As a next step, the {trained} {overprovisioned} network} 
is passed to the \textit{System Optimiser} to be customised for the target use-case (Sec.~\ref{sec:dse}).
At this stage, the \textit{On-device Profiler} performs a number of runs on the target platform and measures the \mbox{per-layer} latency and memory of the overprovisioned CNN.  
Next, {the \textit{SDF Modelling} module} converts the early-exit network to \tool{}'s internal representation (Sec.~\ref{sec:sdf}) and the optimiser traverses the design space following a hardware-aware strategy to generate the highest-performing~design. 



\vspace{-0.1cm}
\section{Early-Exit Network Design}
Given a CNN, the design space of early-exit variants is formed by the free parameters that would yield the resulting early-exit network (Fig.~\ref{InfEngine_v3.pdf}). These include 1)~the number and 2)~positions of 
exits, 
3)~the exit policy, 4)~the training scheme and 5)~the architecture of each exit. In this respect, \tool{} adopts a training strategy that enables the co-optimisation of the number and positioning of early exits, the efficient exploration of various design points and the rapid customisation to the performance goals and target platform.

\vspace{-0.2cm}
\subsection{Number and Placement of Early Exits}
\label{sec:exit_placement}
\vspace{-0.05cm}

The number and positions of early exits have a direct impact on both the accuracy and latency of the end network~\cite{Li2019a}. Nevertheless, so far the conventional exit placement approach for early-exit networks~\cite{branchynet2016,Huang2017,scan2019neurips,sdn_icml_2019,deebert2020acl,spinn2020mobicom} is in a uniform, network- and platform-agnostic manner that severely constrains the optimisation opportunities for the target application. To enable fine-grained customisation and capture a wide range of designs, our placement scheme 1)~allows early-exit positioning along the depth of the CNN, 2)~operates at a fine granularity by allowing exits within building blocks\footnote{For residual and Inception networks, the building blocks are the residual and Inception modules respectively. For networks without skip or multi-path connections (\textit{e.g.} VGG), classifiers can be placed after conv and pool layers.} of the corresponding network family and 3) sets no \textit{a priori} constraint to the number of exits. This approach differs to existing 
schemes~\cite{branchynet2016,Huang2017,scan2019neurips,sdn_icml_2019,deebert2020acl} which keep a coarse granularity, with exits allowed only after a network's building blocks.

With this formulation, the structure of the early-exit model comprises a subset of the candidate early exits. The number and positions of the exits are selected during the DSE \blue{(described in Sec.~\ref{sec:dse})}. Given a subset of early exits, the latency of executing each subgraph on the target platform is known at design time, based on the on-board measurements from the \textit{On-device Profiler} and the developed performance model (detailed in Sec. \ref{sec:perf_model}). As a result, the end-to-end latency of the early-exit network is estimated 
without the need for 
time-consuming on-device runs in the DSE loop. 

\vspace{-0.1cm}

\subsection{Exit Policy} 
\label{sec:exit_strategy}
\vspace{-0.05cm}


\tool{} employs the confidence of each early classifier to identify potentially misclassified samples. At run time, low-confidence outputs are propagated to the next exit to maximise the probability of obtaining an accurate prediction. To estimate the confidence of a prediction, we employ the top-1 output of an exit, \textit{i.e.} $\text{top1}(\boldsymbol{p})$ where $\boldsymbol{p}$ is the output of the softmax layer~\cite{Guo2017}. In this respect, a prediction is considered confident, and thus exits at the i-th classifier, when the condition $\text{top1}(\boldsymbol{p}_i) \ge c^{\text{thr}}$ is satisfied, 
where $c^{\text{thr}}$ is the confidence threshold. If none of the instantiated classifiers exceeds the confidence threshold, the output of the most confident classifier is selected, leading to the following early-exit strategy: $\hat{y}= \max\limits_{i \in [1,N_{\text{exit}}]} \text{top1}(\boldsymbol{p}_i)$,
where $\hat{y}$ is the final output of the network for the current input. 
In our early-exiting scheme, we treat $c^{\text{thr}}$ as a parameter that is shared across early exits and is autotuned by \tool{} to meet the user-specified accuracy and latency. 
The selection of $c^{\text{thr}}$ is exposed to the DSE (Sec.~\ref{sec:dse}), and is co-optimised along with the number and positioning of the intermediate exits. \steve{Shall we say sth here about the fact that this is a simplification to make a point and it's perfectly possible to have a threshold per exit?}

At run time, the \textit{Exit Decision Unit} (see Fig.~\ref{InfEngine_v3.pdf}) considers the latency budget and, using the hardware-aware latency estimator, configures the network to execute up to the last classifier that does not violate the latency constraint. In contrast to existing progressive inference systems \cite{branchynet2016,scan2019neurips,sdn_icml_2019}, whose exit strategy is design-time configurable, \tool{}'s approach enables the generated network to adapt upon deployment and change early-exit policy at run time, based on the device load or specific app requirements.



\vspace{-0.1cm}
\subsection{Early-Exit Training Scheme}
\label{sec:training}
\vspace{-0.1cm}

\noindent
There are two different training schemes that one can follow:  

\textbf{End-to-end training:}
Once the early-classifier positions have been fixed, the network can be trained from scratch, jointly optimising all the classifiers. However, this approach comes at a cost: a multi-objective cost function has to be defined so as to 
balance learning among all classifiers \cite{branchynet2016,Huang2017,sdn_icml_2019}; the classifiers can affect each other's accuracy based on their positioning in the CNN; the network needs new hyperparameter tuning for training; the network might not converge; high turnover time for exploring different exit positions, due to the required retraining and the associated long training time. 
%
On the contrary, a benefit of the end-to-end training is the higher accuracy 
if the classifiers are positioned correctly \cite{sdn_icml_2019}. 

\textbf{\blue{Early-exit-only} training:}
A more modular approach to training early-exit networks is to first train the original network and then the intermediate exits.
Specifically, the network is initially trained with only the last classifier attached. 
Then, intermediate exits are added at all candidate points and trained with the main backbone of the network frozen.\footnote{Weights of ``frozen'' layers do not get updated during the backpropagation phase.}
Last, only the most relevant classifiers can remain attached to the network, depending on their accuracy, exit rate and position in the network.

We select the latter approach as our 
training method in \tool{}, due to its high flexibility in post-training customisation with respect to use-case requirements and target hardware. The first approach of joint 
training as a strict prerequisite to assess a design's performance not only limits the tractability of evaluating many alternative early-exit designs, but also imposes a maintenance cost for deploying such a model in the wild, where it will run on heterogeneous hardware~\cite{embench_2019,facebook2019}. In this case, the overhead of retraining a network variant whenever a different platform is targeted can be prohibitive. 


\vspace{-0.2cm}
\subsection{Early-Exit Architecture}
\label{sec:exit_arch}
\vspace{-0.05cm}

In this work, we treat the exit's architecture 
as an invariant \blue{across the exits}, borrowing the structure of MSDNet classifiers~\cite{Huang2017}.

%% file: sdf_modelling.tex


{
Several deep learning systems~\cite{fpgaconvnet_2018,dnnvm2019tcad,astra_tool_2019} and frameworks~\cite{tensorflow_2016,latte2016pldi,tvm2018osdi} model CNNs as computation graphs.
Typically, the primary goal of this approach is to capture the dependencies between operations and expose their computational and memory requirements in order to apply compiler or hardware optimisations. While this approach suits the execution predictability~\cite{gandiva2018osdi,netadapt2018eccv,wang2019tcad,astra_tool_2019} of typical CNN workloads where the \textit{exact same} computation graph is executed for all inputs, early-exit networks pose a unique challenge: due to their input-dependent early-exit mechanism, samples processed by early-exit models can exit at different points along the network based on their complexity, leading to non-deterministic execution. 
}

\begin{figure}[t]
    \centering
    \vspace{-0.25cm}
    {\includegraphics[width=1\columnwidth,trim={1cm 6.5cm 6.5cm 3cm},clip]{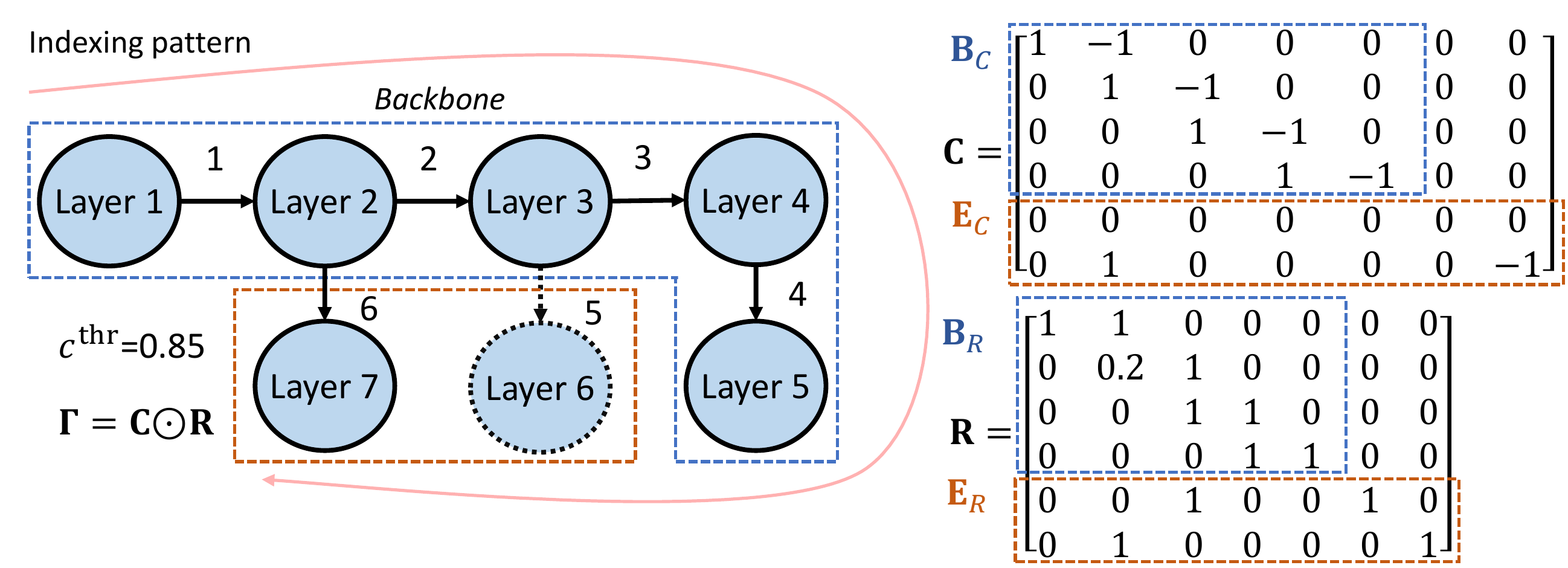}}
    \vspace{-0.7cm}
    \caption{\tool{} design point as an SDF graph.}
    \vspace{-0.5cm}
    \label{fig:sdf_example}
\end{figure}

{
To analyse and optimise the deployment of early-exit networks, an \textit{execution-rate}-driven modelling paradigm is introduced. The proposed modelling framework builds upon synchronous dataflow (SDF)~\cite{lee1987synchronous} and enhances it to capture the unique properties of early-exit CNN workloads.
\tool{} represents \textit{design points} as SDF graphs (SDFGs) that correspond to different early-exit variants (Fig.~\ref{fig:sdf_example}). Given a CNN's overprovisioned architecture, an SDFG, $G=(V,E)$, is formed by assigning one SDF node $v \in V$ to each layer. Its edges $e \in E$ represent data flow between the network's layers.
The SDFG can be represented compactly by a \textit{topology matrix}, $\boldsymbol{\Gamma}$. Each column of $\boldsymbol{\Gamma}$ corresponds to a node and each row to an edge of the SDFG. Each element $\gamma_{ij}$ is an integer value that captures the \textit{production/consumption rate} of node $j$ on edge $i$ and its sign indicates the direction of the data flow.
}

{
The proposed framework enhances the SDF model with two extensions:
1)~The decomposition of the topology matrix ($\mathbf{\Gamma}$) into two matrices ($\mathbf{C}$ and $\mathbf{R}$, Eq.~ (\ref{eq:topology_mat})).
Each of the two matrices allows us to analyse a design point based on the distinct components that affect its performance;
2)~ A method for propagating the effects of local tunings to the overall performance of the design. The proposed approach automatically propagates the effect of a local change to the rest of the SDF graph and calculates the execution rates of different parts of the early-exit network.
}

{
\textbf{Topology Matrix Structural Decomposition.}
To expose the factors that shape the performance of a design point, we decompose the topology matrix into the Hadamard product\footnote{The Hadamard product, denoted by $\odot$, is defined as the elementwise multiplication between two matrices.} between two matrices.
The first matrix is the \textit{connectivity matrix}, denoted by $\mathbf{C}$. Each element $c_{ij} \in \{-1,0,1\}$ indicates whether node $j$ is connected to another node via edge $i$, with 1 and -1 signifying data production and consumption respectively, and 0 no connection. The second matrix is the \textit{rates matrix}, denoted by $\mathbf{R}$. Each element $r_{ij} \in [0,1]$ captures the expected normalised rate of data production or consumption of node $j$ on edge $i$. A value of 0 indicates no data flow and 1 indicates that data are produced or consumed by node $j$ on edge $i$ at \textit{every} input sample. Following this decomposition, for a network with $N_b$ backbone layers and $N$ candidate exit positions, the topology matrix of the SDFG is expressed as follows:
\begin{equation}
    \mathbf{\Gamma} = \mathbf{C} \odot \mathbf{R}
    \label{eq:topology_mat}
\end{equation}
where $\mathbf{\Gamma} \in \mathbb{R}^{|E| \times |V|}$, $\mathbf{C} \in \{-1,0,1\}^{|E| \times |V|}$, $\mathbf{R} \in [0,1]^{|E| \times |V|}$ with $|V| = N_b+N$ nodes and $|E| = N_b+N-1$ edges. To accommodate the real-valued rates matrix $\mathbf{R}$, we extend the conventional SDF and allow the topology matrix to contain real values.
{The two-matrix representation allows us to decouple the architecture of the early-exit network, \textit{i.e.} the number and position of exits, through matrix $\mathbf{C}$, and the impact of the early-exit policy and the inter-exit dynamics on execution rates through matrix $\mathbf{R}$}.
}

{
Fig.~\ref{fig:sdf_example} shows the translation of an example early-exit network to the corresponding SDF graph. In this scenario, the early-exit network consists of \blue{seven} layers, \blue{five} in the backbone architecture ($N_b$=5), two potential early-exit positions ($N$=2) and one selected early-exit ($N_{\text{exit}}$=1). A sample early-exits at the first exit (layer 7) if the prediction confidence exceeds the threshold $c^{\text{thr}}$$=$$0.85$ of the early-exit policy. In this example, we assume that the selected confidence threshold leads to 80\% of the inputs to stop at exit 1.
The 2nd column of matrix $\mathbf{C}$ corresponds to layer 2 and has two edges (2 and 6) to layer 3 and 7 respectively. With the exit rate at exit 1 being 80\%, only 20\% of the inputs carry on from layer 2 to 3 and hence the associated element $r_{2,2}$ of $\mathbf{R}$ is set to 0.2. Finally, the 5-th row of $\mathbf{C}$ is set to zero as the second exit (layer 6) is \mbox{not instantiated}.
}

{
Following our indexing scheme for nodes and edges (red arrow in Fig.~\ref{fig:sdf_example}), the topology matrix $\mathbf{\Gamma}$ and both its constituent matrices $\mathbf{C}$ and $\mathbf{R}$ have a profound structure (Eq.~(\ref{eq:partition_mat})): i) the first $N_{b}$-1 rows capture the backbone CNN architecture forming the \textit{backbone} \mbox{\textit{submatrix} $\mathbf{B}$}. \steve{make sure that we refer elsewhere to the initial network as backbone network. Can we annotate this on Figure 3?} The submatrix is upper bidiagonal with nonzero elements only along the main diagonal and the diagonal above it;\footnote{In the case of multi-branch modules such as residual, Inception and depthwise-separable blocks, the same partitioned structure exists, but $\mathbf{B}$ is not upper bidiagonal.} ii) the rest of the rows are equal to the number of candidate early exits, forming the \textit{exits submatrix} $\mathbf{E}$. Given a selection of number and position of exits, only the corresponding entries of $\mathbf{E}$ are nonzero. Eq.~(\ref{eq:partition_mat}) shows the partitioned structure of $\mathbf{\Gamma}$.
\begin{equation}
    \mathbf{\Gamma} = \left[\begin{array}{ccc}
         \mathbf{B} & | & \mathbf{O}_{(N_{b}-1) \times N}\\
         \hline
         \multicolumn{3}{c}{\mathbf{E}}
    \end{array}\right]
    \label{eq:partition_mat}
\end{equation}
with $\mathbf{B} \in \mathbb{R}^{(N_{b}-1) \times N_{b}}$, $\mathbf{E} \in \mathbb{R}^{N \times (N_{b}+N)}$ and $\mathbf{O}$ is the zero matrix. The same structure is present in $\mathbf{C}$ and $\mathbf{R}$, consisting of the respective submatrices $\mathbf{B}_C$, $\mathbf{E}_C$, $\mathbf{B}_R$ and $\mathbf{E}_R$. {This partitioned structure enables the efficient manipulation of the SDF model by operating only on specific submatrices, as detailed in Sec.~\ref{sec:search_space}.}
}

{
\textbf{Automatic Execution Rate Propagation.}
Given its intrinsic input-dependent data flow, a key characteristic of an early-exit network is the varying execution rate of different parts of the architecture due to its conditional execution. In our modelling framework, we introduce a method to automatically obtain the execution rates of different parts of the network, while propagating them along the architecture.
In conventional SDF theory~\cite{lee1987synchronous}, by solving $\mathbf{\Gamma}\boldsymbol{q}=\mathbf{O}$, we can derive an admissible execution schedule for the topology matrix. In this case, vector $\boldsymbol{q} \in \mathbb{R}^{|V|}$ indicates how many times each node should be executed in one schedule period in order to avoid deadlocks and unbounded buffering. 
}

{
In \tool{}, we introduce an alternative view, tailored to early-exit networks. Under this view, we interpret $q_i$ as the expected normalised \textit{execution rate} of the i-th node, \textit{i.e.} the probability of executing the i-th layer when processing a sample.
To enable this interpretation, we set a constraint on the range of $\boldsymbol{q}$'s elements so that \mbox{$\boldsymbol{q} \in [0,1]^{|V|}$} and proceed to obtain $\boldsymbol{q}$ by solving \mbox{$\mathbf{\Gamma}\boldsymbol{q}=\mathbf{O}$}. Following this approach, in the example of Fig.~\ref{fig:sdf_example}, the vector would be \mbox{$\boldsymbol{q} = [1, 1, 0.2, 0.2, 0.2, 0, 1]^T$}.
Our method enables two key functions.
First, it provides the reinterpretation of the values of $\boldsymbol{q}$ as the execution rates of the network's layers.
For Fig.~\ref{fig:sdf_example}, these values indicate that the first two layers and the early exit (last element of $\boldsymbol{q}$) would process all inputs, while layers 3 to 5 are expected to process 20\% of the inputs, due to the 80\% that would exit early. 
Second, the effect of a local tuning (\textit{e.g.} a selection of $c^{\text{thr}}$ that led to 80\% exit rate in the first exit) is automatically propagated along the SDFG through the calculation of $\boldsymbol{q}$ (\textit{i.e.} the 20\% production rate of layer 2 is propagated to the execution rate of layers 3-5 through the 0.2 value in the corresponding elements of $\boldsymbol{q}$). As a result, the overall impact of a local change on the design's performance is automatically propagated and calculated rapidly through $\boldsymbol{q}$.
}


%% file: design_space_exploration.tex


{
In \tool{}, the basic mapping of an SDF graph (SDFG) is the early-exit network implementation as illustrated in Fig.~\ref{InfEngine_v3.pdf}. The network architecture is first constructed by mapping each SDFG node to a layer and connecting them according to the connectivity \mbox{matrix $\mathbf{C}$}. Furthermore, the \textit{Exit Decision Unit} is configured using the selected confidence threshold $c^{\text{thr}}$. 
While \tool{}'s highly parametrised early-exit network design provides fine-grained control over customisation, it also leads to an excessively large design space.
}

{
In this context, we exploit the analytical power of our modelling framework to efficiently navigate the design space.\hj{I really like this sentence.}
We visit alternative designs by tuning \tool{}'s design parameters through a set of graph \textit{transformations} that can be directly applied over the SDF model (Sec.~\ref{sec:search_space}). To assess the quality of a design point without continuously accessing the target platform, we build analytical models that provide rapid estimates of the attainable latency and memory footprint (Sec.~\ref{sec:perf_model}).
Overall, these exploration and design evaluation techniques are integrated into \tool{}'s optimiser which solves a multi-objective optimisation formulation of the DSE \mbox{task (Sec.~\ref{sec:optimisation})}.
%
}

\subsection{Early-Exit Engine Search Space}
\label{sec:search_space}

{
Based on its early-exit network parametrisation (Sec.~\ref{sec:prog_infer}), \tool{} defines a particular design space formed by 1)~the number of early exits, 2)~their positions along the network and 3)~the early-exit policy. In this respect, we model the configuration of an early-exit design with a tuple of the form $\langle N_{\text{exit}}, \boldsymbol{p}_{\text{exit}}, c^{\text{thr}}, \mathbf{\Gamma}\rangle$, where $N_{\text{exit}}$ is the number of selected exits, $\boldsymbol{p}_{\text{exit}} \in \{0,1\}^{N}$ the positioning vector with the i-th element set to 1 if an exit is placed at position $i$, $c^{\text{thr}}$ the threshold of the early-exit policy, and $\mathbf{\Gamma}$ the topology matrix. 
}

{
Our SDF-based modelling allows us to express the complete design space captured by \tool{} by defining graph transformations for the manipulation of SDFGs.
In this way, any design tuning that transforms the SDFG can be applied directly to the topology matrix $\mathbf{\Gamma}$ by means of efficient algebraic operations. 
\tool{} employs the following set of transformations:
}

{
\begin{enumerate}[leftmargin=0.5cm]
    \setlength{\parskip}{0pt}
    \setlength{\itemsep}{0pt plus 1pt}
    \item \textbf{Early-Exit Repositioning} \textit{exitrepos}($N_{\text{exit}}$,$\boldsymbol{p}_{\text{exit}}$):
The first transformation changes the number and position of early exits along the network by adding and removing early-exit nodes on the SDF graph. Early-exit repositioning modifies both the structure of the SDF graph by altering the architecture of the CNN and the exiting rates as different combinations of early exits have varying early-exit dynamics. As a result, this transformation affects both the connectivity matrix $\mathbf{C}$ and rates matrix $\mathbf{R}$. 
\item \textbf{Confidence-Threshold Tuning} \textit{conftune}($c^{\text{thr}}$):
The second transformation modifies the early-exit policy by tuning the confidence threshold $c^{\text{thr}}$. In particular, low values lead to a less restrictive policy with more samples exiting at the earlier stages of the CNN, while higher values form a more conservative policy with more samples exiting deeper in the network. As a result, a change in $c^{\text{thr}}$ has an impact on the exit rate of each exit and hence affects only the rates matrix $\mathbf{R}$. Since the network architecture remains unchanged, matrix $\mathbf{C}$ is not modified.
\end{enumerate}
}



\noindent
{
Given these transformations, we define the transformation set as $\mathcal{T}=\{exitrepos(N_{\text{exit}},\boldsymbol{p}_{\text{exit}}), conftune(c^{\text{thr}})\}$. To generate a new design point, we apply one or multiple transformations from $\mathcal{T}$ over the current design point $s$:
$s' \xlongleftarrow{t} s, ~~~ t \in \mathcal{T}$.
Formally, the overall search space defined by \tool{} is captured by means of a set $\mathcal{S}$ that contains all reachable alternative designs:
\begin{equation}
    \label{eq:search_space}
    \mathcal{S} = \left\{ s ~|~ s = \left< s_{\text{overprv}}, \mathcal{T}^* \right>  \right\}, ~~~ \mathcal{T}^* \subset \mathcal{T}
\end{equation}
where $s_{\text{overprv}}$ is the overprovisioned variant of the CNN, $\mathcal{T}^*$ is the subset of transformations that are applied on $s_{\text{overprv}}$ to obtain $s$. 
}

{
Our SDF-based framework allows us to express these transformations through algebraic operations directly applied on the topology matrix as described by Algorithm~\ref{alg:transform}. 
The algorithm takes as inputs the $\boldsymbol{\Gamma}$ matrix of the given SDFG and the transformation, $t$, to be applied.
The connectivity matrix $\mathbf{C}$ is affected by the early-exit repositioning (lines 1-6), while the rates matrix $\mathbf{R}$ is affected by both transformations (lines 7-13).
On line 3, a positioning matrix is constructed with $\boldsymbol{p}_{\text{exit}}$ along its diagonal and it is used to left-multiply matrix $\mathbf{E}_C^{\text{all}} \in \mathbb{R}^{N \times (N_b+N)}$, which holds \textit{all} the candidate exits. With this operation, only the rows of $\mathbf{E}_C^{\text{all}}$ that map to the edges between the \textit{selected} exits and the backbone network are selected, with the rest set to zero.
As changes in the number and position of exits do not affect the backbone architecture, submatrix $\mathbf{B}_C$ is not altered and the updated connectivity matrix $\mathbf{C}'$ is produced following Eq. (\ref{eq:partition_mat}) (line 5).
A similar procedure is followed for $\mathbf{R}'$ on lines 7-13. First, the exit rate of each exit is calculated using an efficient memoisation scheme (line 9), detailed in Sec.~\ref{sec:optimisation}.
Next, the exit rates are projected to the associated layer position (line 10). Finally, the production rates of nodes that are connected to exits are updated (line 11) and $\mathbf{R}'$ is formed. As a final step, the updated topology matrix $\mathbf{\Gamma}'$ is constructed (line 14).
}

{
}

\SetArgSty{textnormal} 
\setlength{\textfloatsep}{0pt}
\begin{algorithm}[!t]	
		\footnotesize
		\SetAlgoLined
		\LinesNumbered
		\DontPrintSemicolon
		\KwIn{Topology matrix $\mathbf{\Gamma}=\mathbf{C} \odot \mathbf{R}$}
		\nonl
		\myinput{Transformation $t \in \mathcal{T}$}
		\KwOut{Updated topology matrix $\mathbf{\Gamma}'$} 
		
		\texttt{/*} - - - \textit{update connectivity matrix $\mathbf{C}$} - - - \texttt{*/}\;
		\If{$t$ \textbf{is} \textit{exitrepos}($N_{\text{exit}}$, $\boldsymbol{p}_{\text{exit}}$)}{
    		$\mathbf{E}_{\text{sel}} \leftarrow \textit{diag}(\boldsymbol{p}_{\text{exit}})$ // \textit{Form the positioning matrix} \;
    		$\mathbf{E}_C' \leftarrow \mathbf{E}_{\text{sel}} \mathbf{E}_C^{\text{all}}$ // \textit{Update early-exit submatrix $\mathbf{E}_C$}
            \hspace*{20em}
            \rlap{\smash{$\left.\begin{array}{@{}c@{}}\\{}\\{}\\{}\\{}\end{array}\color{black}\right\}%
            \color{black}\begin{tabular}{l}Backbone submatrix $\mathbf{B}_C$\\  is not affected by changes \\
            in the no. and position of exits.
            \end{tabular}$}}\;
    		$\mathbf{C'} \leftarrow \text{UpdateMatrix}(\mathbf{B}_C, \mathbf{E}_C')$ \;
		}
		\nonl\;
		
		\texttt{/*} - - - \textit{update rates matrix $\mathbf{R}$} - - - \texttt{*/}\;
		\If{$t$ \textbf{is} \textit{exitrepos}($N_{\text{exit}}$, $\boldsymbol{p}_{\text{exit}}$) \textbf{or} \textit{conftune}($c^{\text{thr}}$)}{
            $\boldsymbol{r}_{\text{exit}} \leftarrow \text{MemoisedData}(N_{\text{exit}}, \boldsymbol{p}_{\text{exit}}, c^{\text{thr}})$ // \textit{Obtain exit rates through memoisation (Sec.~\ref{sec:optimisation})} \;
    		$\boldsymbol{r}_{\text{layer}} = \mathbf{E'}_C(:,1:N_b)^T \boldsymbol{r}_{\text{exit}}$ // \textit{Map exit rates to their layer positions} \;
    		$\mathbf{B}_R' = \mathbf{B}_R \odot \textit{diag}(\boldsymbol{r}_{\text{exit}})$ // \textit{Update the backbone submatrix} \; 
    		$\mathbf{R}' \leftarrow \text{UpdateMatrix}(\mathbf{B}_R', \mathbf{E}_R)$ \;
		}
		
		$\mathbf{\Gamma}' = \mathbf{C}' \odot \mathbf{R}'$ \texttt{//}  \textit{reconstruct topology matrix} \;
		
		\caption{\small Design tuning as algebraic operations}
		\label{alg:transform}
\end{algorithm}


\subsection{Performance and Memory Footprint Model}
\label{sec:perf_model}

To estimate the latency and memory footprint of each design point, we developed an analytical performance model that leverages \tool{}'s modelling framework.
As a first step, after the given CNN is augmented with exits at all candidate positions, the \textit{On-device Profiler} executes a number of on-board benchmark runs to measure the per-layer execution time of the overprovisioned early-exit CNN, denoted by $l_i$ for $\forall i \in [1,|V|]$. The execution time measurements are integrated into vector $\boldsymbol{l} = [l_1, l_2, ..., l_{|V|}]^T$. This phase takes place only \textit{once} upfront and hence the DSE task does not require access to the target platform.
Given the topology matrix $\mathbf{\Gamma}$ of a design point $s=\langle N_{\text{exit}}, \boldsymbol{p}_{\text{exit}}, c^{\text{thr}}, \mathbf{\Gamma}\rangle$, the execution rate vector $\boldsymbol{q}$ is calculated using the automatic execution rate propagation scheme (Sec.~\ref{sec:sdf}). With each element of $\boldsymbol{q}$ giving the expected execution rate of each layer in design point $s$, the hardware-specific average latency of processing an input $I$ can be estimated as $L_{\text{hw}}(I, s) = \boldsymbol{q}^T \boldsymbol{l}$.
For memory consumption, due to the typically small batch size of the inference stage, the model size (\textit{i.e.} the CNN's weights) dominate the run-time memory. In this respect, we define the memory footprint vector $\boldsymbol{m} \in \{0\} \cup \mathbb{Z}^{+|V|}$ with the i-th element holding the footprint of the i-th node's weights. Given vector $\mathds{1}(\boldsymbol{q}$$>$$0) \in \{0,1\}^{|V|}$ masking only the nodes that are used in design point $s$, the memory consumption of $s$ is estimated as 
$m(s) = \mathds{1}(\boldsymbol{q}\text{>}1)^T \boldsymbol{m}$.

\vspace{-0.2cm}
\subsection{System Optimisation}
\vspace{-0.1cm}
\label{sec:optimisation}


To evaluate the quality of the design points that lie within the search space and select the highest-performing ones, we cast the problem as multi-objective optimisation (MOO) and design an objective function that reflects the key requirements of the use-case. With respect to latency, the majority of existing early-exit works \cite{branchynet2016,Huang2017,sdn_icml_2019,spinn2020mobicom} rely on the theoretical FLOPs as a proxy to its real processing speed. Such an approach ignores essential platform-specific characteristics including caching, I/O and hardware-level features, leading to the FLOP count not accurately capturing the actual attainable performance of executing a CNN on a particular processing platform \cite{obj_det_tradeoffs_2017,embench_2019}. 
In contrast, we employ a hardware-aware approach that utilises real device latency, alongside {memory footprint and} accuracy, as metrics to assess the quality of each design and drive \tool{}'s search towards high-performance designs.

In our MOO setup, we employ two objective functions \mbox{(Eq.~(\ref{eq:coopt}, \ref{eq:coopt_w_sla}))} that reduce the multi-objective problem to a single objective by means of the weighted sum and $\epsilon$-constraint methods \cite{marler2004survey} respectively. In the weighted sum formulation, the modelling of the interplay between quality metrics plays a decisive role in shaping the trade-offs to be explored \cite{marler2004survey}; in \tool{} the dynamics between accuracy and latency determine how much additional latency cost we allow to pay for each percentage point (pp) of accuracy gain. 

As a first step, for the weights to closely capture the importance of each metric in the target application \cite{marler2010weighted}, the accuracy and latency of each design point $s$ are divided by the accuracy and latency of the original CNN
respectively, to obtain a non-dimensional objective function. 
\begin{figure}[t]
    \centering
    \vspace{-0.2cm}
    {
    \includegraphics[width=0.85\columnwidth,trim={8.5cm 13.5cm 7cm 13cm},clip]{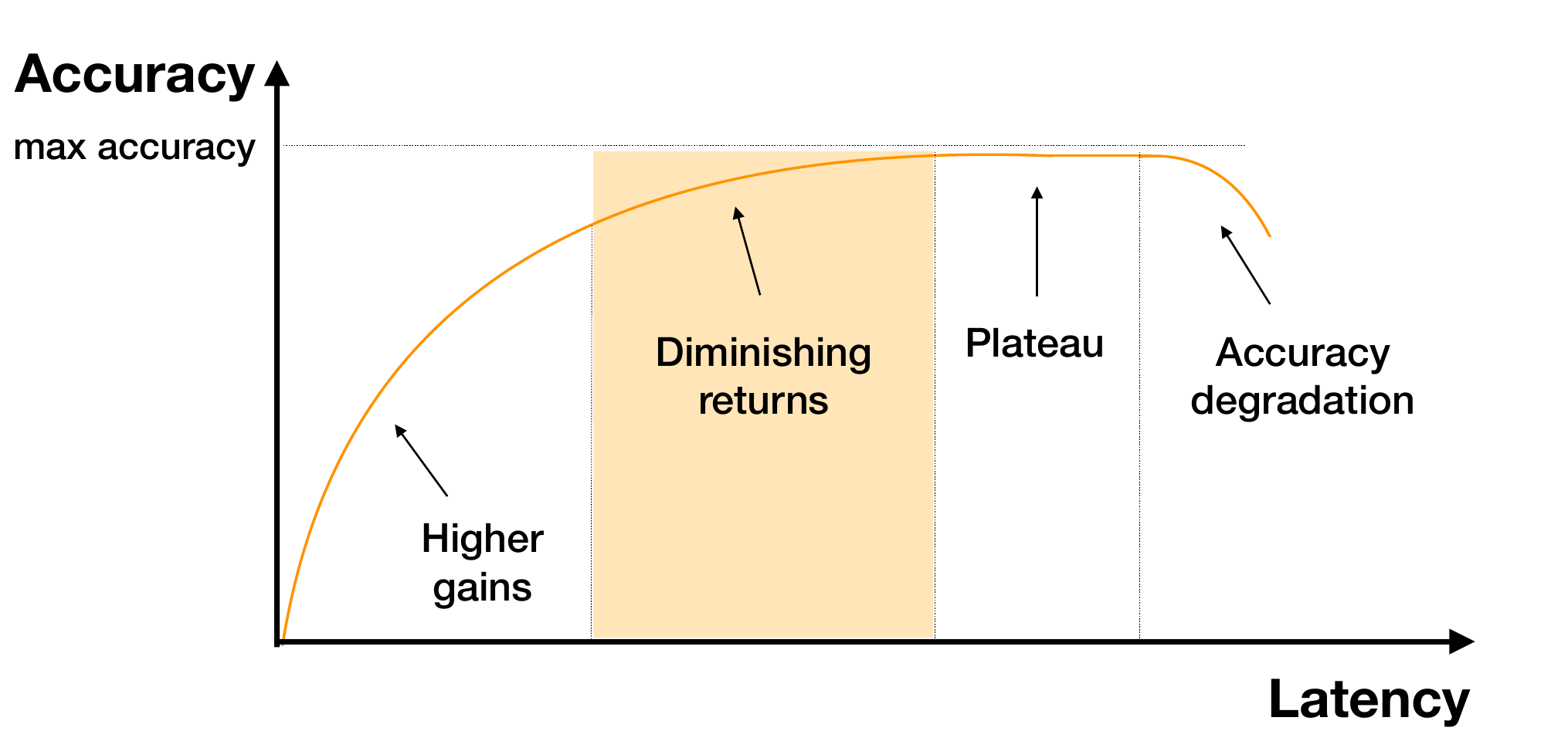}
    }
    \vspace{-1.2em}
    \caption{Accuracy-latency trade-off.}
    \label{Latency_Accuracy_v2.pdf}
\end{figure}
%
%
Next, we model the dynamics of the accuracy-latency trade-off through a non-linear logarithmic function (Fig.~\ref{Latency_Accuracy_v2.pdf}). 
The selected function reflects the fact that the accuracy-latency trade-off is more prominent in the beginning of the network compared to the end, where the accuracy typically plateaus and we obtain diminishing returns on the computation time. In this respect, we set the accuracy weight to 1 and tune the latency weight $\mathrm{w}_{\text{lat}}$ via grid search to obtain the most beneficial trade-off in the DSE phase, aiming for a solution in the highlighted area of Fig.~\ref{Latency_Accuracy_v2.pdf}. 
Overall, we pose the following MOO problems:
\vspace{-0.1cm}
\vspace{-0.1cm}
\begingroup
\begin{align}
    & \max\limits_{s}~\frac{A(s)}{A^{\text{max}}} -  \mathrm{w}_{\text{lat}} \cdot \log\left(\frac{L_{\text{hw}}(I, s)}{L_{\text{hw}}^{\text{max}}} + 1\right) 
    \label{eq:coopt}
    \\ & \text{ s.t. } L_{\text{hw}}(I, s) \le \epsilon \quad \& \quad m(s) \le m_{\text{max}} 
    \label{eq:coopt_w_sla}
\end{align}
\endgroup
where $A(s)$ is the average accuracy of the current early-exit design $s$, $L_{\text{hw}}(I,s)$ and $m(s)$ are the latency and memory footprint of $s$ on the target platform respectively, and $\epsilon$ and $m_{\text{max}}$ are the user-specified upper bound on latency and maximum memory capacity of the platform respectively. The objective functions aim to either: \mbox{1)~co-optimise} accuracy and latency (Eq.~(\ref{eq:coopt})) or 2)~also impose latency and memory constraints (Eq.~(\ref{eq:coopt},\ref{eq:coopt_w_sla})).

{
\textbf{Efficient Evaluation through Memoisation.}
Given a CNN, the optimisation problems are defined over the set of all points $\mathcal{S}$ in the presented design space (Sec.~\ref{sec:search_space}).
For the objective functions to be evaluated, the exit rate of each exit is required to construct $\boldsymbol{\Gamma}$ (see $\boldsymbol{r}_{\text{exit}}$ in Algorithm~\ref{alg:transform}) and then calculate $L_{\text{hw}}(I,s)$ (Sec.~\ref{sec:perf_model}), together with the accuracy $A(s)$. 
Typically, to obtain these values, the design point $s$ 
would have to be materialised in the form of a CNN and run over the calibration set, monitoring how many samples stopped at each exit together with whether they were classified correctly.
This process leads to the excessive overhead of running inference over the calibration set for each examined design point.}

{
To alleviate this high cost, we exploit the key observation that by processing each sample of the calibration set \textit{once} using the overprovisioned CNN and storing only 1) the top-1 value and 2) whether the sample was correctly classified at each exit, we can evaluate the accuracy and exit rates of any design point.
For a calibration set of size $|D|$, $N$ candidate exit positions and $N_{\text{conf}}$ candidate confidence thresholds, the memoised evaluation would require $2|D| \cdot N \cdot N_{\text{conf}}$ elements to be stored, which can be used to evaluate the objective function of any $s$. As an example of the required space, for the validation set of ImageNet ($|D|$$=$$50,000$), ResNet-56 ($N$$=$$58$) and three confidence thresholds ($N_{\text{conf}}$$=$$3$), the storage requirement is 66~MB.  With this approach, given the selection of exits and the confidence threshold of an examined design point, the expensive inference process is replaced with a fast lookup of the associated values from the memoised data and applying the rule of \tool{}'s exit strategy (Sec.~\ref{sec:exit_strategy}). {This process takes place offline at design time and hence places no burden on the end device upon deployment.} 
}

\textbf{Optimiser.}
Given a CNN, the objective functions of the defined optimisation problems can be evaluated for all design points given the introduced memoisation scheme and the performance model of Sec.~\ref{sec:perf_model}. 
To jointly optimise the number and positioning of early exits, we cast them as a search problem where we aim to select adequate early-exit positions that optimise the objective function. 
In this respect, for a CNN with $N$ possible exit positions, we seek the value of the binary positioning vector $\boldsymbol{p}_{\text{exit}} \in \{0,1\}^{N}$ that optimises the target objective function.

In theory, the optimal early-exit design could be obtained by means of exhaustive enumeration. 
Given the different number of exits, exit positions and early-exit policies, the overall number of candidate designs to be examined can be calculated as $N_{\text{conf}} \cdot 2^{N}$$-$$1$ 
where $N_{\text{conf}}$ is the number of distinct examined values for the confidence threshold (\textit{e.g.} \{0.4, 0.6, 0.8\}). 
With an increase in the network's depth, $N$ increases accordingly, and brute-force enumeration quickly becomes 
intractable. To this end, a heuristic optimiser is adopted to obtain a solution in the non-convex space.

In this work, Simulated Annealing (SA)~\cite{Reeves_1993} has been selected as the basis of the developed optimiser. Given the set of SDF transformations $\mathcal{T}$ defined in Sec.~\ref{sec:search_space}, the neighbourhood 
of a design $s$ is defined as the set of design points that can be reached from $s$ by applying one of the operations $t \in \mathcal{T}$. 
Overall, the optimiser navigates the design space by considering the described SDF transformations and converges to a solution 
of the target objective function.
%
%
%
%
To prune the exponential space, we introduce a prior by initially not allowing 
exits to be 
in adjacent positions.
After the optimiser has selected the highest-performing design, \tool{} explores adjacent positions of the already chosen exits, as a refinement step. 



%% file: evaluation.tex
{
In this section, we evaluate \tool{}'s performance against a random search optimiser, the improvement over the state-of-the-art early-exit methods under varying latency budgets and the performance gains over hand-crafted CNN models. 
} 
\vspace{-0.1cm}
\subsection{Experimental Setup}
\label{sec:exp_setup}
\vspace{-0.1cm}

In our experiments, we target two platforms with different resource characteristics (Table \ref{tab:experimental_setup}): a server-grade desktop computer and an Nvidia Jetson Xavier AGX. 
For the latter, we evaluate on two different power profiles (30W, underclocked~10W) by adjusting the \blue{thermal design power (TDP)} and clock rate of the CPU and GPU. We build our framework on top of \texttt{\small PyTorch} (v1.1.0) and \texttt{\small torchvision} (v0.3.0) compiled with Nvidia cuDNN.

\textbf{Benchmarks.}
We show the generalisability of our system across different benchmark networks which vary in terms of depth, computational load and architecture. Specifically, we include \mbox{VGG-16}~\cite{Simonyan14c} as a large and computationally intensive network that has conventional single-layer connectivity; ResNet~\cite{He_2016} and Inception-v3~\cite{Szegedy_2017} as representative mainstream networks from the residual and Inception-based network families, that include non-trivial connectivity via the residual and Inception blocks respectively. We also compare \tool{} with two hand-optimised networks: the state-of-the-art early-exit network MSDNet~\cite{Huang2017}, and MobileNetV2~\cite{mobilenetv2}, a highly-optimised architecture for resource-constrained devices. 


\begin{table}[t]
    \centering
    \vspace{-0.3cm}
    \caption{\small {Target Platforms}}
    \vspace{-0.3cm}
    \setlength{\tabcolsep}{2pt}
    \resizebox{\linewidth}{!}{
        \scriptsize
        \begin{tabular}{l l l l l l}
            \toprule
            \textbf{Platform} & \textbf{Processor} & \textbf{Memory} & \textbf{GPU} & \textbf{TDP} \\
            \midrule
            \scriptsize
            \begin{tabular}[l]{@{}l@{}} Server\\ \phantom{(8 cores, HT)} \end{tabular} &
            \scriptsize
            \begin{tabular}[l]{@{}l@{}} Intel i7-7820X \\ (8 cores, HT) \end{tabular} &
            \scriptsize
            \begin{tabular}[l]{@{}c@{}} 128GB DDR4 \\ @ 2133MHz \end{tabular} &
            \scriptsize
            \begin{tabular}[l]{@{}l@{}} Nvidia GTX \\ 1080Ti \end{tabular} &
            \scriptsize
            \begin{tabular}[l]{@{}l@{}} 400W \\ \phantom{(8 cores, HT)} \end{tabular} \\
            
            \scriptsize
            \begin{tabular}[l]{@{}l@{}} Jetson Xavier  \end{tabular} &
            \scriptsize
            \begin{tabular}[l]{@{}l@{}} 8-core ARM-Karmel v8.2 \end{tabular} &
            \scriptsize
            \begin{tabular}[l]{@{}l@{}} 16GB LPDDR4x \end{tabular} &
            \scriptsize
            \begin{tabular}[l]{@{}l@{}} 512-core Volta \end{tabular} &
            \scriptsize
            \begin{tabular}[l]{@{}l@{}} 30W, (u)10W \end{tabular} \\
            
            \bottomrule
        \end{tabular}
    }
    \label{tab:experimental_setup}
\end{table}

\textbf{Datasets and Training Scheme.}
We evaluate the effectiveness of our approach on the CIFAR-100 \cite{Krizhevsky2009} and \mbox{ImageNet~\cite{Fei-Fei2010}} image classification datasets. 
We use the process described in each model's implementation for data augmentation and preprocessing, such as scaling and cropping the input, stochastic horizontal flipping and channel colour normalisation.
%
In \tool{}'s early-exit-only training policy (Sec. \ref{sec:training}), for the initial step of training the main network, we train our own networks for CIFAR-100 using the authors' guidelines for hyperparameter selection. For ImageNet, we use the pretrained networks distributed by \texttt{\small torchvision}, while for MSDNet, we train the ImageNet variant from \cite{Huang2017}. 
To train the early exits in the second step, we continue for an additional 300 and 90 epochs for CIFAR-100 and ImageNet respectively, using the same batch size and an Adam optimiser, with momentum $0.9$ and weight \mbox{decay $10^{-4}$}.

\vspace{-0.2cm}
\subsection{Evaluation of Proposed Optimiser}
\vspace{-0.1cm}

\begin{figure}[t]
    \vspace{-0.4cm}
    \centering
    \includegraphics[width=0.725\columnwidth]{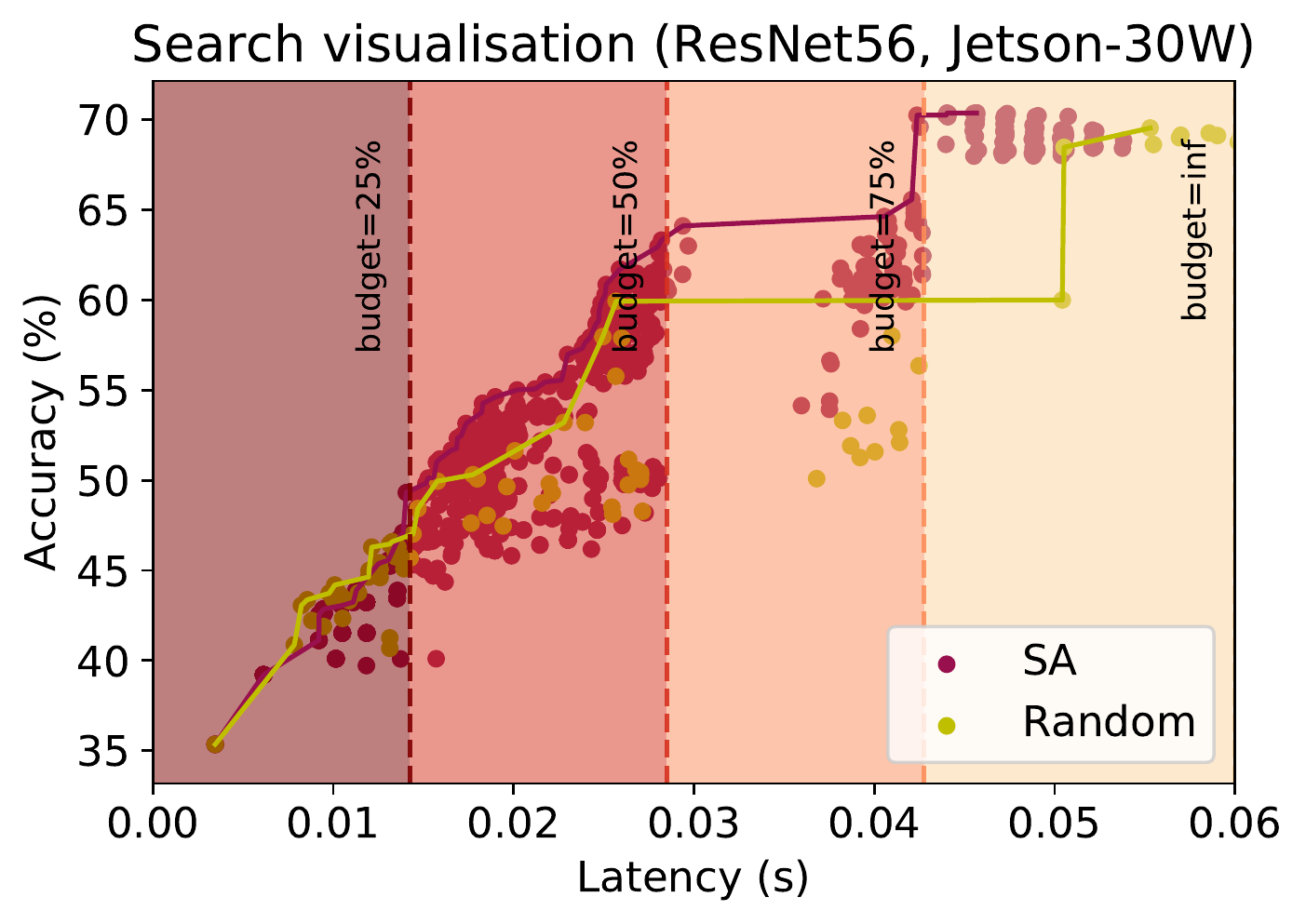}
    \vspace{-0.3cm}
    \caption{Visualisation of explored design space.}
    \label{fig:search_space}
    \vspace{0.1cm}
\end{figure}

To evaluate our DSE, we compare our algorithm with a random search (RS) baseline. Specifically, we compare each exploration of the search space \textit{under the same runtime budget}. We employ \mbox{ResNet-56} on CIFAR-100 targeting the 30-watt AGX, across four settings by varying the latency SLA.
Fig.~\ref{fig:search_space} visualises the points visited by each search, clustered by SLA deadline. Across latency budgets, our SA-based optimiser 
yields a Pareto front with designs that dominate the RS Pareto points, achieving 3.39 and 10.32 percentage points (pp) higher accuracy under 28- and 42-ms SLAs respectively.
We also observe that RS tends to \textit{revisit} already examined designs due to remembering nothing but the best examined design, leading to inefficient utilisation of the available runtime with fewer distinct design points examined. 
We note that RS has found marginally better designs in the beginning of the 25\% SLA due to the small acceptable search space \blue{ caused by the latter exits becoming infeasible as they violate the tight latency deadline.
}


\vspace{-0.2cm}
\subsection{Evaluation against Early-Exit Frameworks}
\vspace{-0.05cm}

In this section, we evaluate \tool{} against the state-of-the-art early-exit frameworks, namely BranchyNet and SDN. 
{BranchyNet~\cite{branchynet2016} uses two manually placed early exits and an entropy-based exit policy. We place two early exits at 33\% and 66\% of FLOPs and perform a sweep over entropy thresholds to tune the value for each experiment.
For SDN~\cite{sdn_icml_2019}, we place 6 early exits equidistantly with respect to FLOPs and perform a sweep over confidence thresholds to adjust the exit policy for each experiment.}

\begin{figure}[t]
    \centering
    \vspace{-0.4cm}
    \small
    \begin{subfigure}{0.48\columnwidth}
      \centering
      \includegraphics[width=\columnwidth]{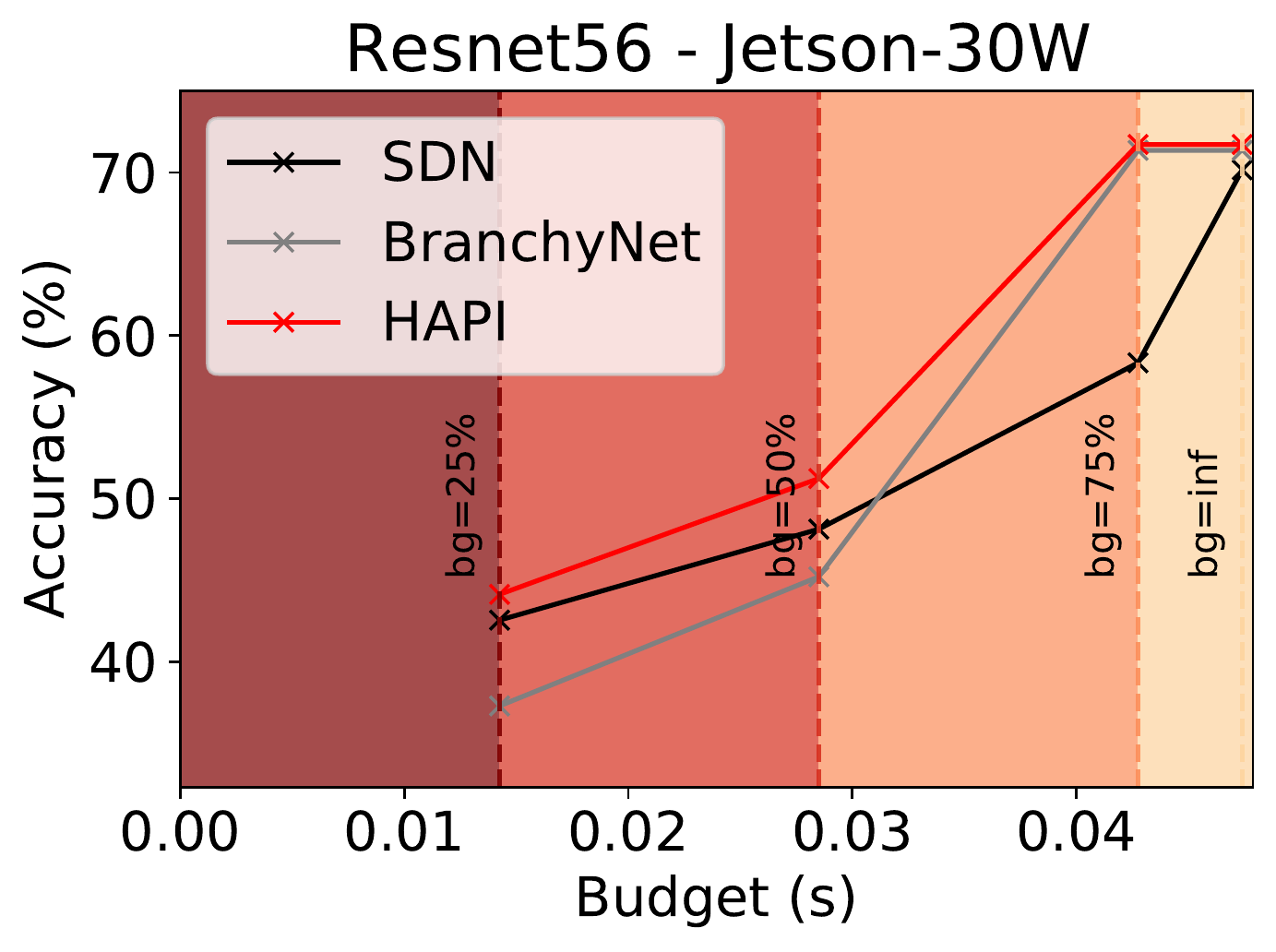}
      \caption{ResNet-56 on Jetson (30W).}
      \label{fig:resnet56-jetson30}
    \end{subfigure}%
    \hfill
    \begin{subfigure}{0.48\columnwidth}
      \centering
      \includegraphics[width=\columnwidth]{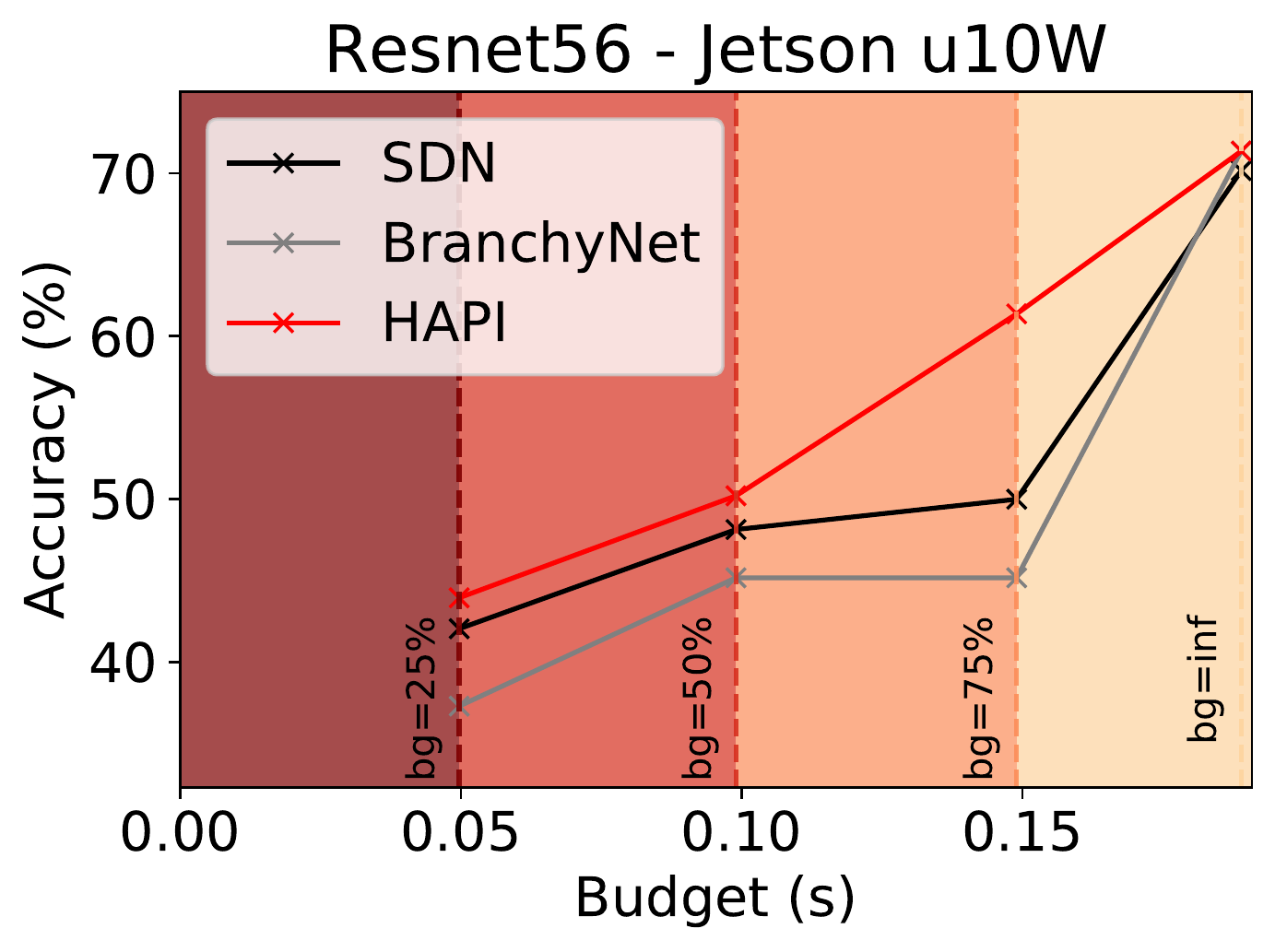}
      \caption{ResNet-56 on Jetson (u10W).}
      \label{fig:resnet56-jetson5}
    \end{subfigure}
    \hfill
    \begin{subfigure}{0.48\columnwidth}
      \centering
      \includegraphics[width=\columnwidth]{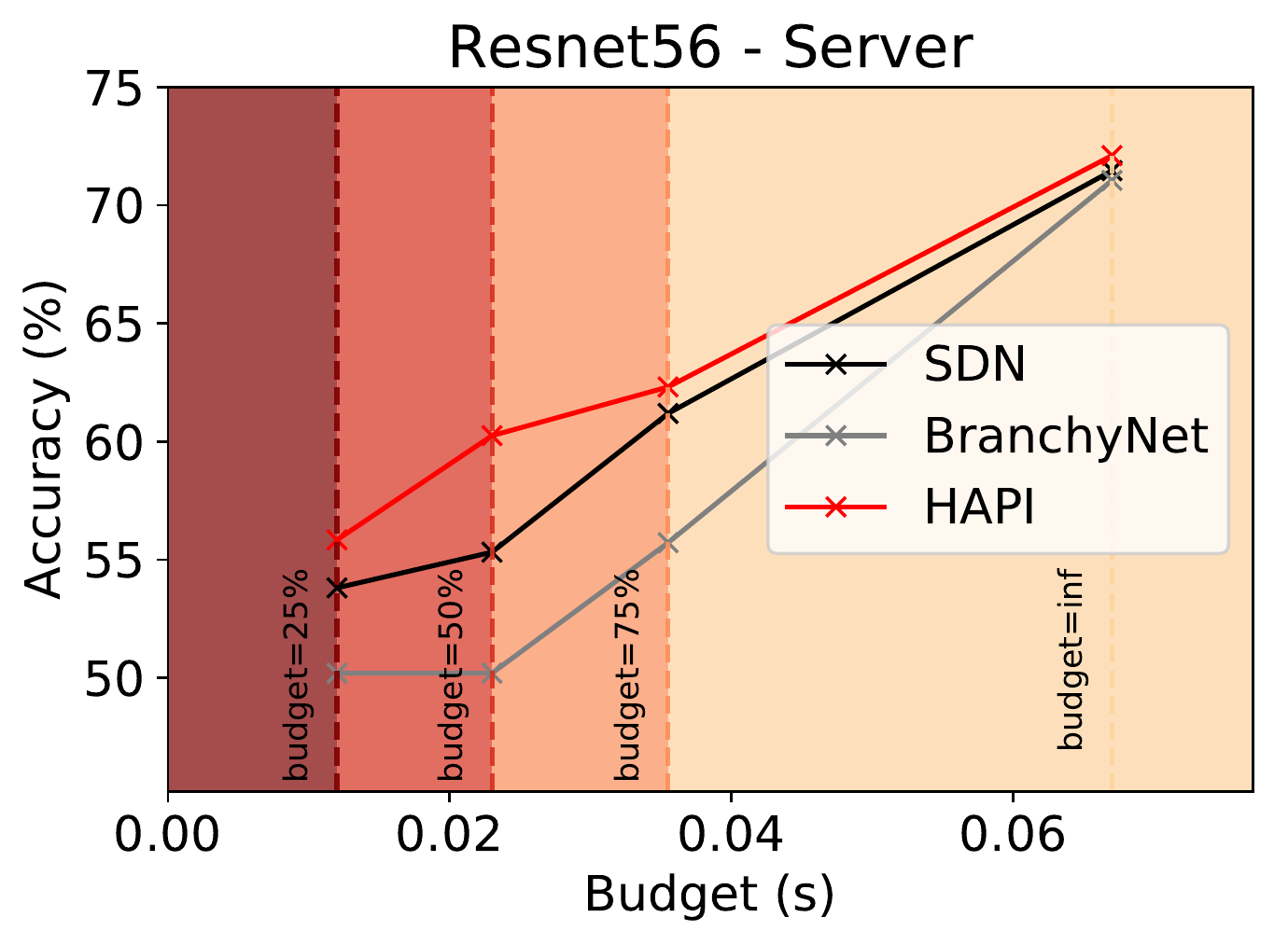}
      \caption{ResNet-56 on server.}
      \label{fig:resnet56-server}
    \end{subfigure}
    \hfill
    \begin{subfigure}{0.48\columnwidth}
      \centering
      \includegraphics[width=\columnwidth]{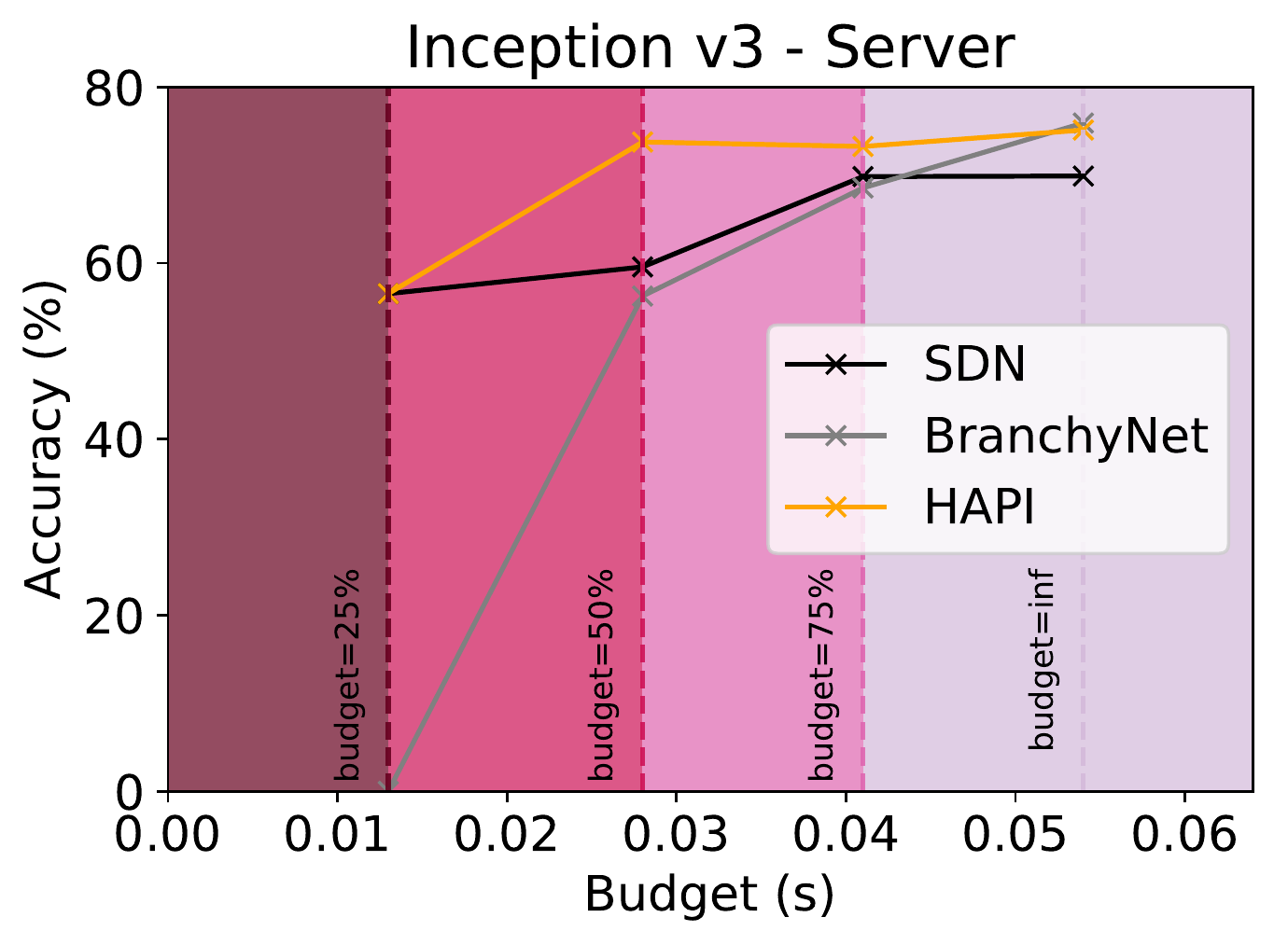}
      \caption{Inception-v3 on server.}
      \label{fig:inception-server}
    \end{subfigure}
    \hfill
    \begin{subfigure}{0.48\columnwidth}
      \includegraphics[width=\columnwidth]{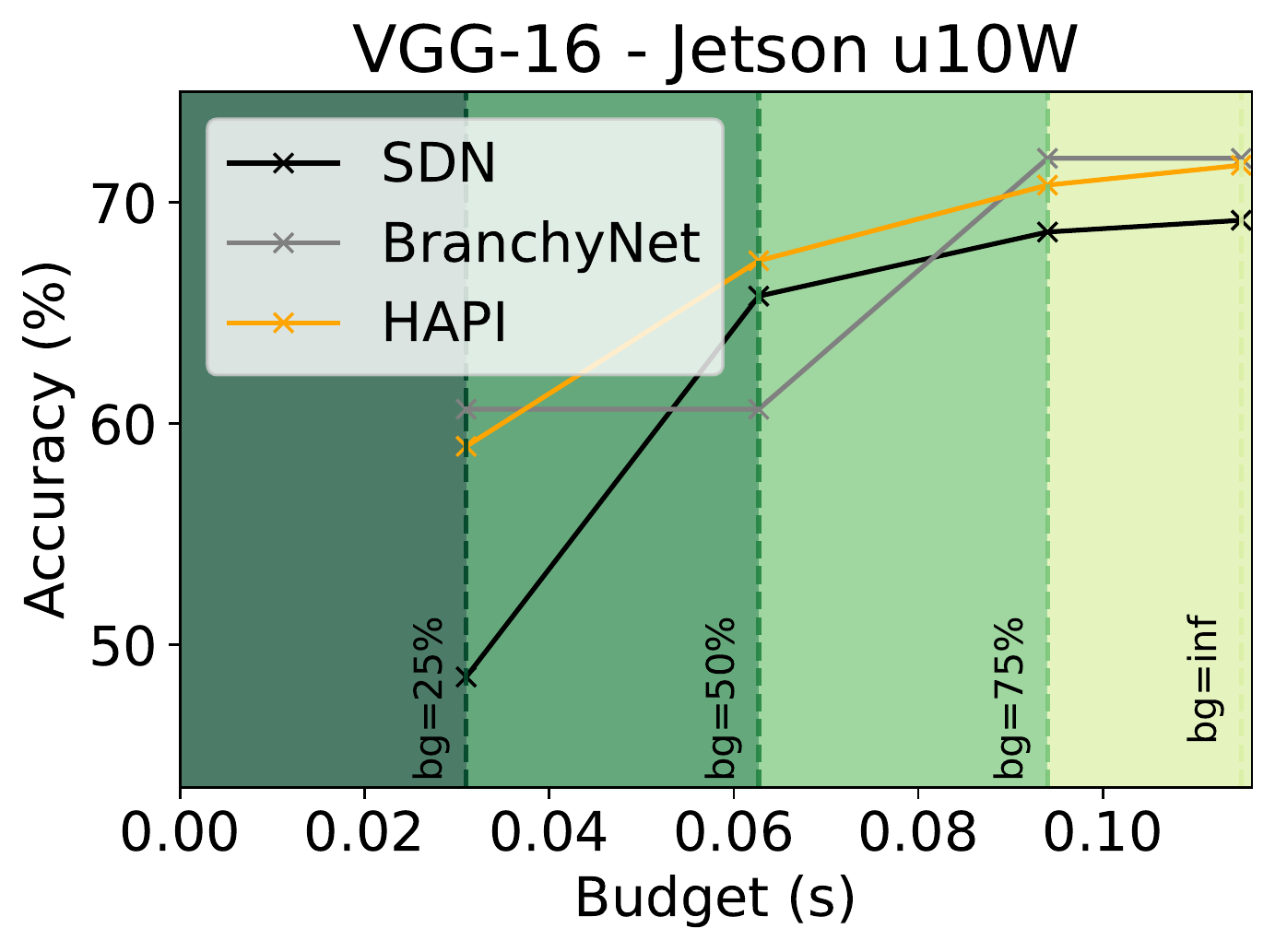}
      \caption{VGG-16 on Jetson (u10W).}
      \label{fig:vgg-jetson5}
    \end{subfigure}
    \hfill
    \begin{subfigure}{0.48\columnwidth}
      \includegraphics[width=\columnwidth]{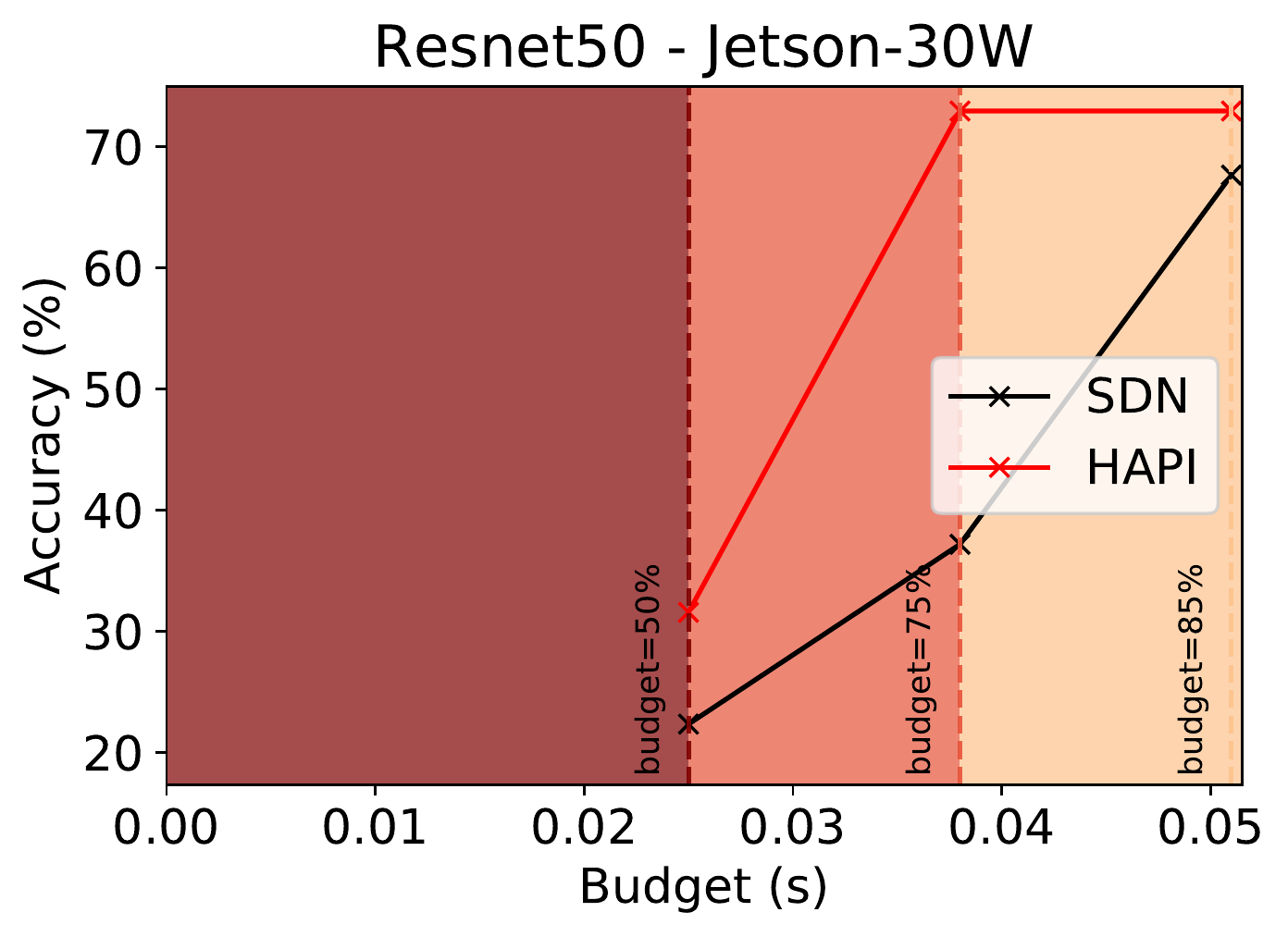}
      \caption{ResNet-50 on Jetson (30W).}
      \label{fig:Resnet50_Jetson-30W}
    \end{subfigure}
    \vspace{-0.65em}
    \caption{Comparison of \tool{} with SDN and BranchyNet.\stelios{Annotate datasets somehow.}}
    \label{fig:baselines}
\end{figure}

Fig.~\ref{fig:resnet56-jetson30}-\ref{fig:vgg-jetson5} show the respective early-exit designs under various maximum latency SLAs, represented by the different colour gradients, on CIFAR-100. 
\tool{} generates \textit{consistently} more accurate designs for a variety of given latency budgets, when compared to the other strategies not explicitly optimising for the hardware platform or the SLA deadline.
Specifically, for ResNet-56 (\mbox{Fig.~\ref{fig:resnet56-jetson30}-\ref{fig:resnet56-server}}), our search yields higher-accuracy designs across devices and budgets, ranging from 0.5 to 6 pp gain over SDN and up to 55 pp over BranchyNet. In particular, this situation manifests when the BranchyNet's first exit (statically positioned at 33\% of the network's FLOPs) violates the SLA. 
We further evaluate \tool{} on different architectures, such as Inception-v3 (Fig.~\ref{fig:inception-server}) on the server and VGG-16 (Fig.~\ref{fig:vgg-jetson5}) on the u10W-profile AGX. We observe the same behaviour for BranchyNet in the low-latency SLAs, while \tool{} delivers up to 14.2 pp higher accuracy over SDN for a budget of 30~ms. 


We showcase \tool{}'s scalability by selectively training and optimising ResNet-50 on ImageNet (Fig. \ref{fig:Resnet50_Jetson-30W}). \tool{} dominates SDN's solutions across budgets on the 30W AGX, with accuracy gains of 4.1-35.7 pp (avg. 16.36 pp). At a 38-ms budget, we observe a significant accuracy improvement of 35.7~pp. This is due to the substantial latency overhead of executing early classifiers on the larger-scale ImageNet. Thus, with \tool{} generating a design with fewer exits than SDN's static 6-exit scheme, the CNN can reach deeper layers, without latency violations, and hence achieve higher accuracy.



\vspace{-0.2cm}
\subsection{Comparison with Hand-Crafted Networks}
\label{sec:msdnet}
\vspace{-0.1cm}

In this section, the quality of \tool{} designs is assessed with respect to two state-of-the-art hand-optimised models: i) the early-exit MSDNet and ii) the lightweight MobileNetV2.

\noindent
\textbf{Hand-tuned Early-exit Network.}
This is investigated on CIFAR-100 by comparing the achieved performance in the accuracy-latency space. 
Our MSDNet model comprises 10 exits, each positioned after a block.
We treat MSDNet as a network pre-populated with all candidate exits and for each latency budget we let \tool{} generate the highest-performing subset of exits and the associated $c^{\text{thr}}$ value.



\begin{figure}[t]
    \vspace{-0.4cm}
    \centering
    \includegraphics[width=\columnwidth]{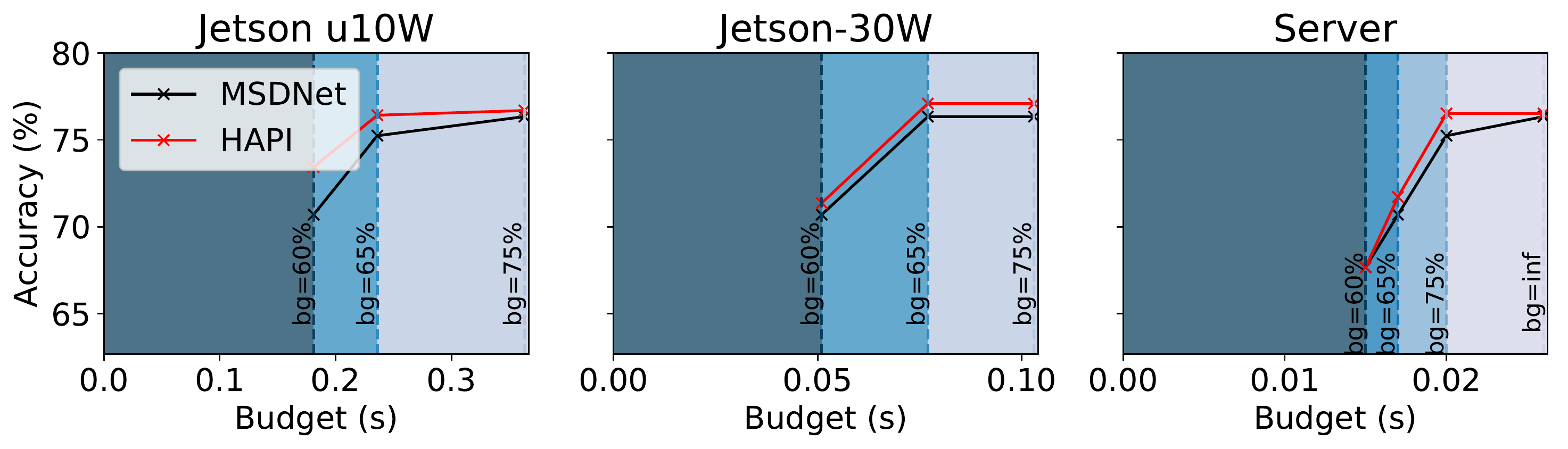}
    \vspace{-0.5cm}
    \caption{Comparison of \tool{} with MSDNet-CIFAR.}
    \vspace{-0.2cm}
    \label{fig:cifar-baselines}
\end{figure}

As shown in Fig. \ref{fig:cifar-baselines}, our framework is able to sustain the performance of MSDNet across all settings, while achieving higher accuracy under certain cases. On a severely power-constrained device (Fig. \ref{fig:cifar-baselines} left), \tool{} yields up to 2.74 pp of accuracy improvement with a latency constraint of less than 200 ms, with an average gain of 1.42 pp across the different latency budgets. In the 30W mode of Jetson AGX (Fig. \ref{fig:cifar-baselines} middle), \tool{} achieves up to 0.75 pp under a 80-ms latency constraint. Finally, in the case of the server-grade platform (Fig. \ref{fig:cifar-baselines} right), \tool{} yields up to 1.28 pp over MSDNet. 

In the case of ImageNet, 
\tool{} selected the fully populated network. This 
can be attributed to most of the computations of MSDNet for ImageNet being located in the model's backbone. Thus, selecting a subset of exits does not significantly benefit latency, but has a non-negligible impact on accuracy.
%
\blue{With respect to deployability, MSDNet's computationally heavy architecture struggles to meet stringent requirements on resource-constrained platforms. On 30W AGX, \tool{}'s ResNet-56 
achieves similar accuracy to MSDNet at 41~ms, yielding 20\% speedup over MSDNet's 50~ms. For even tighter constraints, MSDNet does not contain any viable exit. } 


\begin{figure}[t]
    \vspace{-0.3cm}
    \centering
    \includegraphics[width=\columnwidth,trim={0.1cm 0.25cm 0.1cm 0.25cm},clip]{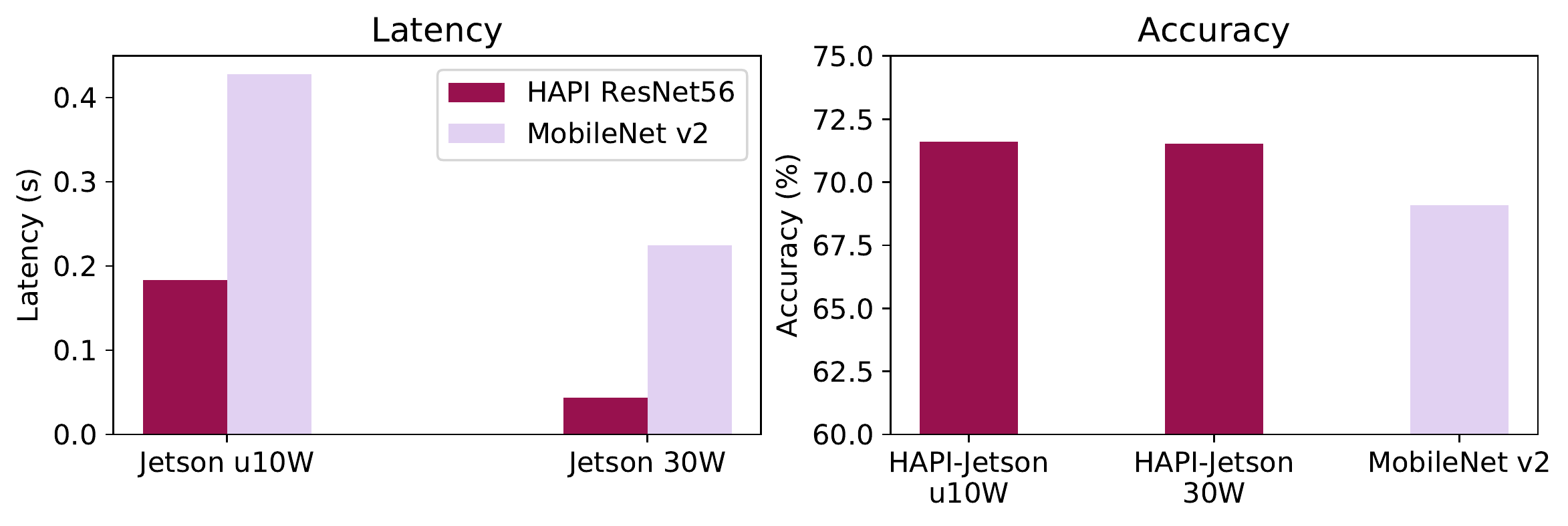}
    \vspace{-0.6cm}
    \caption{Comparison of \tool{}-ResNet with \mbox{MobileNetV2}.}
    \label{fig:mobilenetv2}
\end{figure}


\noindent
\textbf{Hand-tuned Lightweight Network.}
\label{sec:mobilenetv2}
%
Although we pose \tool{} as an orthogonal, model-agnostic methodology to architecture-specific techniques, we compare with the state-of-the-art lightweight MobileNetV2, taking its end latency as our budget for optimisation. 
As shown in Fig.~\ref{fig:mobilenetv2}, \tool{} outperforms MobileNetV2 on Jetson with an accuracy gain of 2.53 and 2.45 pp and a speedup of 2.33$\times$ and 5.11$\times$ under the u10- and 30-watt profiles respectively.

%% file: main.bbl

\begin{thebibliography}{43}


\ifx \showCODEN    \undefined \def \showCODEN     #1{\unskip}     \fi
\ifx \showDOI      \undefined \def \showDOI       #1{#1}\fi
\ifx \showISBNx    \undefined \def \showISBNx     #1{\unskip}     \fi
\ifx \showISBNxiii \undefined \def \showISBNxiii  #1{\unskip}     \fi
\ifx \showISSN     \undefined \def \showISSN      #1{\unskip}     \fi
\ifx \showLCCN     \undefined \def \showLCCN      #1{\unskip}     \fi
\ifx \shownote     \undefined \def \shownote      #1{#1}          \fi
\ifx \showarticletitle \undefined \def \showarticletitle #1{#1}   \fi
\ifx \showURL      \undefined \def \showURL       {\relax}        \fi
\providecommand\bibfield[2]{#2}
\providecommand\bibinfo[2]{#2}
\providecommand\natexlab[1]{#1}
\providecommand\showeprint[2][]{arXiv:#2}

\bibitem[\protect\citeauthoryear{Abadi et~al\mbox{.}}{Abadi
  et~al\mbox{.}}{2016}]%
        {tensorflow_2016}
\bibfield{author}{\bibinfo{person}{Mart\'{\i}n Abadi} {et~al\mbox{.}}}
  \bibinfo{year}{2016}\natexlab{}.
\newblock \showarticletitle{{TensorFlow: A System for Large-scale Machine
  Learning}}. In \bibinfo{booktitle}{\emph{Proceedings of the 12th USENIX
  Conference on Operating Systems Design and Implementation (OSDI)}}.
  \bibinfo{pages}{265--283}.
\newblock


\bibitem[\protect\citeauthoryear{Almeida, Laskaridis, Leontiadis, Venieris, and
  Lane}{Almeida et~al\mbox{.}}{2019}]%
        {embench_2019}
\bibfield{author}{\bibinfo{person}{Mario Almeida}, \bibinfo{person}{Stefanos
  Laskaridis}, \bibinfo{person}{Ilias Leontiadis},
  \bibinfo{person}{Stylianos~I. Venieris}, {and} \bibinfo{person}{Nicholas~D.
  Lane}.} \bibinfo{year}{2019}\natexlab{}.
\newblock \showarticletitle{{EmBench: Quantifying Performance Variations of
  Deep Neural Networks Across Modern Commodity Devices}}. In
  \bibinfo{booktitle}{\emph{International Workshop on Embedded and Mobile Deep
  Learning (EMDL)}}.
\newblock


\bibitem[\protect\citeauthoryear{Chen et~al\mbox{.}}{Chen
  et~al\mbox{.}}{2018}]%
        {tvm2018osdi}
\bibfield{author}{\bibinfo{person}{Tianqi Chen} {et~al\mbox{.}}}
  \bibinfo{year}{2018}\natexlab{}.
\newblock \showarticletitle{{TVM: An Automated End-to-End Optimizing Compiler
  for Deep Learning}}. In \bibinfo{booktitle}{\emph{13th {USENIX} Symposium on
  Operating Systems Design and Implementation ({OSDI})}}.
\newblock


\bibitem[\protect\citeauthoryear{Fei-Fei, Deng, and Li}{Fei-Fei
  et~al\mbox{.}}{2010}]%
        {Fei-Fei2010}
\bibfield{author}{\bibinfo{person}{L. Fei-Fei}, \bibinfo{person}{J. Deng},
  {and} \bibinfo{person}{K. Li}.} \bibinfo{year}{2010}\natexlab{}.
\newblock \showarticletitle{{ImageNet: Constructing a Large-Scale Image
  Database}}.
\newblock \bibinfo{journal}{\emph{Journal of Vision}} (\bibinfo{year}{2010}).
\newblock
\showISBNx{978-1-4244-3992-8}
\showISSN{1534-7362}


\bibitem[\protect\citeauthoryear{Guo, Pleiss, Sun, and Weinberger}{Guo
  et~al\mbox{.}}{2017}]%
        {Guo2017}
\bibfield{author}{\bibinfo{person}{Chuan Guo}, \bibinfo{person}{Geoff Pleiss},
  \bibinfo{person}{Yu Sun}, {and} \bibinfo{person}{Kilian~Q. Weinberger}.}
  \bibinfo{year}{2017}\natexlab{}.
\newblock \showarticletitle{{On Calibration of Modern Neural Networks}}. In
  \bibinfo{booktitle}{\emph{International Conference on Machine Learning}}.
\newblock


\bibitem[\protect\citeauthoryear{Han, Shen, Philipose, Agarwal, Wolman, and
  Krishnamurthy}{Han et~al\mbox{.}}{2016}]%
        {mcdnn_2016}
\bibfield{author}{\bibinfo{person}{Seungyeop Han}, \bibinfo{person}{Haichen
  Shen}, \bibinfo{person}{Matthai Philipose}, \bibinfo{person}{Sharad Agarwal},
  \bibinfo{person}{Alec Wolman}, {and} \bibinfo{person}{Arvind Krishnamurthy}.}
  \bibinfo{year}{2016}\natexlab{}.
\newblock \showarticletitle{{MCDNN}: An Approximation-Based Execution Framework
  for Deep Stream Processing Under Resource Constraints}. In
  \bibinfo{booktitle}{\emph{Proceedings of the 14th Annual International
  Conference on Mobile Systems, Applications, and Services (MobiSys)}}.
\newblock


\bibitem[\protect\citeauthoryear{{He} et~al\mbox{.}}{{He}
  et~al\mbox{.}}{2018}]%
        {maskrcnn2018tpami}
\bibfield{author}{\bibinfo{person}{K. {He}} {et~al\mbox{.}}}
  \bibinfo{year}{2018}\natexlab{}.
\newblock \showarticletitle{{Mask R-CNN}}.
\newblock \bibinfo{journal}{\emph{IEEE Transactions on Pattern Analysis and
  Machine Intelligence (TPAMI)}} (\bibinfo{year}{2018}).
\newblock


\bibitem[\protect\citeauthoryear{He, Zhang, Ren, and Sun}{He
  et~al\mbox{.}}{2016}]%
        {He_2016}
\bibfield{author}{\bibinfo{person}{K. He}, \bibinfo{person}{X. Zhang},
  \bibinfo{person}{S. Ren}, {and} \bibinfo{person}{J. Sun}.}
  \bibinfo{year}{2016}\natexlab{}.
\newblock \showarticletitle{{Deep Residual Learning for Image Recognition}}. In
  \bibinfo{booktitle}{\emph{IEEE Conference on Computer Vision and Pattern
  Recognition (CVPR)}}.
\newblock


\bibitem[\protect\citeauthoryear{Hsieh et~al\mbox{.}}{Hsieh
  et~al\mbox{.}}{2018}]%
        {focus_2018}
\bibfield{author}{\bibinfo{person}{Kevin Hsieh} {et~al\mbox{.}}}
  \bibinfo{year}{2018}\natexlab{}.
\newblock \showarticletitle{{Focus: Querying Large Video Datasets with Low
  Latency and Low Cost}}. In \bibinfo{booktitle}{\emph{13th USENIX Conference
  on Operating Systems Design and Implementation (OSDI)}}.
\newblock


\bibitem[\protect\citeauthoryear{Huang et~al\mbox{.}}{Huang
  et~al\mbox{.}}{2018}]%
        {Huang2017}
\bibfield{author}{\bibinfo{person}{Gao Huang} {et~al\mbox{.}}}
  \bibinfo{year}{2018}\natexlab{}.
\newblock \showarticletitle{{Multi-Scale Dense Networks for Resource Efficient
  Image Classification}}. In \bibinfo{booktitle}{\emph{International Conference
  on Learning Representations (ICLR)}}.
\newblock


\bibitem[\protect\citeauthoryear{{Huang} et~al\mbox{.}}{{Huang}
  et~al\mbox{.}}{2017}]%
        {obj_det_tradeoffs_2017}
\bibfield{author}{\bibinfo{person}{J. {Huang}} {et~al\mbox{.}}}
  \bibinfo{year}{2017}\natexlab{}.
\newblock \showarticletitle{{Speed/Accuracy Trade-Offs for Modern Convolutional
  Object Detectors}}. In \bibinfo{booktitle}{\emph{IEEE Conference on Computer
  Vision and Pattern Recognition (CVPR)}}.
\newblock


\bibitem[\protect\citeauthoryear{Kaya, Hong, and Dumitras}{Kaya
  et~al\mbox{.}}{2019}]%
        {sdn_icml_2019}
\bibfield{author}{\bibinfo{person}{Yigitcan Kaya}, \bibinfo{person}{Sanghyun
  Hong}, {and} \bibinfo{person}{Tudor Dumitras}.}
  \bibinfo{year}{2019}\natexlab{}.
\newblock \showarticletitle{{Shallow-Deep Networks: Understanding and
  Mitigating Network Overthinking}}. In \bibinfo{booktitle}{\emph{International
  Conference on Machine Learning (ICML)}}.
\newblock


\bibitem[\protect\citeauthoryear{{Kouris} and {Bouganis}}{{Kouris} and
  {Bouganis}}{2018}]%
        {kouris2018iros}
\bibfield{author}{\bibinfo{person}{A. {Kouris}} {and} \bibinfo{person}{C.
  {Bouganis}}.} \bibinfo{year}{2018}\natexlab{}.
\newblock \showarticletitle{{Learning to Fly by MySelf: A Self-Supervised
  CNN-Based Approach for Autonomous Navigation}}. In
  \bibinfo{booktitle}{\emph{2018 IEEE/RSJ International Conference on
  Intelligent Robots and Systems (IROS)}}.
\newblock


\bibitem[\protect\citeauthoryear{{Kouris}, {Venieris}, and {Bouganis}}{{Kouris}
  et~al\mbox{.}}{2020}]%
        {cascadecnn2020date}
\bibfield{author}{\bibinfo{person}{A. {Kouris}}, \bibinfo{person}{S.~I.
  {Venieris}}, {and} \bibinfo{person}{C. {Bouganis}}.}
  \bibinfo{year}{2020}\natexlab{}.
\newblock \showarticletitle{{A Throughput-Latency Co-Optimised Cascade of
  Convolutional Neural Network Classifiers}}. In \bibinfo{booktitle}{\emph{2020
  Design, Automation Test in Europe Conference Exhibition (DATE)}}.
  \bibinfo{pages}{1656--1661}.
\newblock


\bibitem[\protect\citeauthoryear{Kouris, Venieris, and Bouganis}{Kouris
  et~al\mbox{.}}{2018}]%
        {cascadecnn2018}
\bibfield{author}{\bibinfo{person}{A. Kouris}, \bibinfo{person}{S.~I.
  Venieris}, {and} \bibinfo{person}{C.~S. Bouganis}.}
  \bibinfo{year}{2018}\natexlab{}.
\newblock \showarticletitle{{CascadeCNN}: Pushing the Performance Limits of
  Quantisation in Convolutional Neural Networks}. In
  \bibinfo{booktitle}{\emph{28th International Conference on Field Programmable
  Logic and Applications (FPL)}}.
\newblock
\showISSN{1946-1488}


\bibitem[\protect\citeauthoryear{Krizhevsky}{Krizhevsky}{2009}]%
        {Krizhevsky2009}
\bibfield{author}{\bibinfo{person}{Alex Krizhevsky}.}
  \bibinfo{year}{2009}\natexlab{}.
\newblock \bibinfo{booktitle}{\emph{Learning multiple layers of features from
  tiny images}}.
\newblock \bibinfo{type}{{T}echnical {R}eport}.
\newblock


\bibitem[\protect\citeauthoryear{Laskaridis, Venieris, Almeida, Leontiadis, and
  Lane}{Laskaridis et~al\mbox{.}}{2020}]%
        {spinn2020mobicom}
\bibfield{author}{\bibinfo{person}{Stefanos Laskaridis},
  \bibinfo{person}{Stylianos~I. Venieris}, \bibinfo{person}{Mario Almeida},
  \bibinfo{person}{Ilias Leontiadis}, {and} \bibinfo{person}{Nicholas~D.
  Lane}.} \bibinfo{year}{2020}\natexlab{}.
\newblock \showarticletitle{{SPINN: Synergistic Progressive Inference of Neural
  Networks over Device and Cloud}}. In \bibinfo{booktitle}{\emph{The 26th
  Annual International Conference on Mobile Computing and Networking
  (MobiCom)}}.
\newblock


\bibitem[\protect\citeauthoryear{Lee and Messerschmitt}{Lee and
  Messerschmitt}{1987}]%
        {lee1987synchronous}
\bibfield{author}{\bibinfo{person}{Edward~A Lee} {and} \bibinfo{person}{David~G
  Messerschmitt}.} \bibinfo{year}{1987}\natexlab{}.
\newblock \showarticletitle{{Synchronous Data Flow}}.
\newblock \bibinfo{journal}{\emph{Proc. IEEE}} \bibinfo{volume}{75},
  \bibinfo{number}{9} (\bibinfo{year}{1987}), \bibinfo{pages}{1235--1245}.
\newblock


\bibitem[\protect\citeauthoryear{Lee, Ajanthan, and Torr}{Lee
  et~al\mbox{.}}{2019a}]%
        {lee2019snip}
\bibfield{author}{\bibinfo{person}{Namhoon Lee},
  \bibinfo{person}{Thalaiyasingam Ajanthan}, {and} \bibinfo{person}{Philip
  Torr}.} \bibinfo{year}{2019}\natexlab{a}.
\newblock \showarticletitle{{SNIP}: {Single}-{Shot} {Network} {Pruning} {based}
  {on} {Connection} {Sensitivity}}. In \bibinfo{booktitle}{\emph{International
  Conference on Learning Representations (ICLR)}}.
\newblock


\bibitem[\protect\citeauthoryear{Lee, Venieris, Dudziak, Bhattacharya, and
  Lane}{Lee et~al\mbox{.}}{2019b}]%
        {Lee_2019}
\bibfield{author}{\bibinfo{person}{Royson Lee}, \bibinfo{person}{Stylianos~I.
  Venieris}, \bibinfo{person}{Lukasz Dudziak}, \bibinfo{person}{Sourav
  Bhattacharya}, {and} \bibinfo{person}{Nicholas~D. Lane}.}
  \bibinfo{year}{2019}\natexlab{b}.
\newblock \showarticletitle{{MobiSR: Efficient On-Device Super-Resolution
  Through Heterogeneous Mobile Processors}}. In \bibinfo{booktitle}{\emph{The
  25th Annual International Conference on Mobile Computing and Networking
  (MobiCom)}}.
\newblock


\bibitem[\protect\citeauthoryear{Li, Zhang, Qi, Yang, and Huang}{Li
  et~al\mbox{.}}{2019}]%
        {Li2019a}
\bibfield{author}{\bibinfo{person}{Hao Li}, \bibinfo{person}{Hong Zhang},
  \bibinfo{person}{Xiaojuan Qi}, \bibinfo{person}{Ruigang Yang}, {and}
  \bibinfo{person}{Gao Huang}.} \bibinfo{year}{2019}\natexlab{}.
\newblock \showarticletitle{{Improved Techniques for Training Adaptive Deep
  Networks}}. In \bibinfo{booktitle}{\emph{IEEE International Conference on
  Computer Vision (ICCV)}}.
\newblock


\bibitem[\protect\citeauthoryear{Marler and Arora}{Marler and Arora}{2004}]%
        {marler2004survey}
\bibfield{author}{\bibinfo{person}{R~Timothy Marler} {and}
  \bibinfo{person}{Jasbir~S Arora}.} \bibinfo{year}{2004}\natexlab{}.
\newblock \showarticletitle{Survey of multi-objective optimization methods for
  engineering}.
\newblock \bibinfo{journal}{\emph{Structural and multidisciplinary
  optimization}} (\bibinfo{year}{2004}).
\newblock


\bibitem[\protect\citeauthoryear{Marler and Arora}{Marler and Arora}{2010}]%
        {marler2010weighted}
\bibfield{author}{\bibinfo{person}{R~Timothy Marler} {and}
  \bibinfo{person}{Jasbir~S Arora}.} \bibinfo{year}{2010}\natexlab{}.
\newblock \showarticletitle{The weighted sum method for multi-objective
  optimization: new insights}.
\newblock \bibinfo{journal}{\emph{Structural and multidisciplinary
  optimization}} \bibinfo{volume}{41}, \bibinfo{number}{6}
  (\bibinfo{year}{2010}), \bibinfo{pages}{853--862}.
\newblock


\bibitem[\protect\citeauthoryear{Reeves}{Reeves}{1993}]%
        {Reeves_1993}
\bibfield{editor}{\bibinfo{person}{Colin~R. Reeves}} (Ed.).
  \bibinfo{year}{1993}\natexlab{}.
\newblock \bibinfo{booktitle}{\emph{{Modern Heuristic Techniques for
  Combinatorial Problems}}}.
\newblock \bibinfo{publisher}{John Wiley \& Sons, Inc.}
\newblock


\bibitem[\protect\citeauthoryear{Sandler, Howard, Zhu, Zhmoginov, and
  Chen}{Sandler et~al\mbox{.}}{2018}]%
        {mobilenetv2}
\bibfield{author}{\bibinfo{person}{Mark Sandler}, \bibinfo{person}{Andrew
  Howard}, \bibinfo{person}{Menglong Zhu}, \bibinfo{person}{Andrey Zhmoginov},
  {and} \bibinfo{person}{Liang-Chieh Chen}.} \bibinfo{year}{2018}\natexlab{}.
\newblock \showarticletitle{MobileNetV2: Inverted Residuals and Linear
  Bottlenecks}. In \bibinfo{booktitle}{\emph{IEEE Conference on Computer Vision
  and Pattern Recognition (CVPR)}}.
\newblock


\bibitem[\protect\citeauthoryear{Simonyan and Zisserman}{Simonyan and
  Zisserman}{2015}]%
        {Simonyan14c}
\bibfield{author}{\bibinfo{person}{K. Simonyan} {and} \bibinfo{person}{A.
  Zisserman}.} \bibinfo{year}{2015}\natexlab{}.
\newblock \showarticletitle{{Very Deep Convolutional Networks for Large-Scale
  Image Recognition}}. In \bibinfo{booktitle}{\emph{International Conference on
  Learning Representations (ICLR)}}.
\newblock


\bibitem[\protect\citeauthoryear{Sivathanu, Chugh, Singapuram, and
  Zhou}{Sivathanu et~al\mbox{.}}{2019}]%
        {astra_tool_2019}
\bibfield{author}{\bibinfo{person}{Muthian Sivathanu}, \bibinfo{person}{Tapan
  Chugh}, \bibinfo{person}{Sanjay~S. Singapuram}, {and} \bibinfo{person}{Lidong
  Zhou}.} \bibinfo{year}{2019}\natexlab{}.
\newblock \showarticletitle{{Astra: Exploiting Predictability to Optimize Deep
  Learning}}. In \bibinfo{booktitle}{\emph{Proceedings of the Twenty-Fourth
  International Conference on Architectural Support for Programming Languages
  and Operating Systems (ASPLOS)}}.
\newblock


\bibitem[\protect\citeauthoryear{{Sze}, {Chen}, {Yang}, and {Emer}}{{Sze}
  et~al\mbox{.}}{2017}]%
        {eff_proc_dnns_2017}
\bibfield{author}{\bibinfo{person}{V. {Sze}}, \bibinfo{person}{Y. {Chen}},
  \bibinfo{person}{T. {Yang}}, {and} \bibinfo{person}{J.~S. {Emer}}.}
  \bibinfo{year}{2017}\natexlab{}.
\newblock \showarticletitle{{Efficient Processing of Deep Neural Networks: A
  Tutorial and Survey}}.
\newblock \bibinfo{journal}{\emph{Proc. of the IEEE}} (\bibinfo{year}{2017}).
\newblock


\bibitem[\protect\citeauthoryear{Szegedy, Ioffe, Vanhoucke, and Alemi}{Szegedy
  et~al\mbox{.}}{2017}]%
        {Szegedy_2017}
\bibfield{author}{\bibinfo{person}{Christian Szegedy}, \bibinfo{person}{Sergey
  Ioffe}, \bibinfo{person}{Vincent Vanhoucke}, {and} \bibinfo{person}{Alexander
  Alemi}.} \bibinfo{year}{2017}\natexlab{}.
\newblock \showarticletitle{{Inception-v4, Inception-ResNet and the Impact of
  Residual Connections on Learning}}. In \bibinfo{booktitle}{\emph{AAAI}}.
\newblock


\bibitem[\protect\citeauthoryear{Taylor, Marco, Wolff, Elkhatib, and
  Wang}{Taylor et~al\mbox{.}}{2018}]%
        {adapt_model_sel2018lctes}
\bibfield{author}{\bibinfo{person}{Ben Taylor}, \bibinfo{person}{Vicent~Sanz
  Marco}, \bibinfo{person}{Willy Wolff}, \bibinfo{person}{Yehia Elkhatib},
  {and} \bibinfo{person}{Zheng Wang}.} \bibinfo{year}{2018}\natexlab{}.
\newblock \showarticletitle{{Adaptive Deep Learning Model Selection on Embedded
  Systems}}. In \bibinfo{booktitle}{\emph{Proceedings of the 19th ACM
  SIGPLAN/SIGBED International Conference on Languages, Compilers, and Tools
  for Embedded Systems (LCTES)}}. \bibinfo{pages}{31–43}.
\newblock


\bibitem[\protect\citeauthoryear{Teerapittayanon, McDanel, and
  Kung}{Teerapittayanon et~al\mbox{.}}{2016}]%
        {branchynet2016}
\bibfield{author}{\bibinfo{person}{Surat Teerapittayanon},
  \bibinfo{person}{Bradley McDanel}, {and} \bibinfo{person}{HT Kung}.}
  \bibinfo{year}{2016}\natexlab{}.
\newblock \showarticletitle{{BranchyNet: Fast Inference via Early Exiting from
  Deep Neural Networks}}. In \bibinfo{booktitle}{\emph{International Conference
  on Pattern Recognition (ICPR)}}.
\newblock


\bibitem[\protect\citeauthoryear{Truong, Barik, Totoni, Liu, Markley, Fox, and
  Shpeisman}{Truong et~al\mbox{.}}{2016}]%
        {latte2016pldi}
\bibfield{author}{\bibinfo{person}{Leonard Truong}, \bibinfo{person}{Rajkishore
  Barik}, \bibinfo{person}{Ehsan Totoni}, \bibinfo{person}{Hai Liu},
  \bibinfo{person}{Chick Markley}, \bibinfo{person}{Armando Fox}, {and}
  \bibinfo{person}{Tatiana Shpeisman}.} \bibinfo{year}{2016}\natexlab{}.
\newblock \showarticletitle{{Latte: A Language, Compiler, and Runtime for
  Elegant and Efficient Deep Neural Networks}}. In
  \bibinfo{booktitle}{\emph{Proceedings of the 37th ACM SIGPLAN Conference on
  Programming Language Design and Implementation (PLDI)}}.
\newblock


\bibitem[\protect\citeauthoryear{{Venieris} and {Bouganis}}{{Venieris} and
  {Bouganis}}{2019}]%
        {fpgaconvnet_2018}
\bibfield{author}{\bibinfo{person}{S.~I. {Venieris}} {and} \bibinfo{person}{C.
  {Bouganis}}.} \bibinfo{year}{2019}\natexlab{}.
\newblock \showarticletitle{{fpgaConvNet: Mapping Regular and Irregular
  Convolutional Neural Networks on FPGAs}}.
\newblock \bibinfo{journal}{\emph{IEEE Transactions on Neural Networks and
  Learning Systems (TNNLS)}} \bibinfo{volume}{30}, \bibinfo{number}{2}
  (\bibinfo{year}{2019}), \bibinfo{pages}{326--342}.
\newblock


\bibitem[\protect\citeauthoryear{Wang, Liu, Lin, Lin, and Han}{Wang
  et~al\mbox{.}}{2019}]%
        {wang2019haq}
\bibfield{author}{\bibinfo{person}{Kuan Wang}, \bibinfo{person}{Zhijian Liu},
  \bibinfo{person}{Yujun Lin}, \bibinfo{person}{Ji Lin}, {and}
  \bibinfo{person}{Song Han}.} \bibinfo{year}{2019}\natexlab{}.
\newblock \showarticletitle{{HAQ: Hardware-Aware Automated Quantization with
  Mixed Precision}}. In \bibinfo{booktitle}{\emph{CVPR}}.
\newblock


\bibitem[\protect\citeauthoryear{{Wang}, {Ananthanarayanan}, {Zeng}, {Goel},
  {Pathania}, and {Mitra}}{{Wang} et~al\mbox{.}}{2019}]%
        {wang2019tcad}
\bibfield{author}{\bibinfo{person}{S. {Wang}}, \bibinfo{person}{G.
  {Ananthanarayanan}}, \bibinfo{person}{Y. {Zeng}}, \bibinfo{person}{N.
  {Goel}}, \bibinfo{person}{A. {Pathania}}, {and} \bibinfo{person}{T.
  {Mitra}}.} \bibinfo{year}{2019}\natexlab{}.
\newblock \showarticletitle{{High-Throughput CNN Inference on Embedded ARM
  big.LITTLE Multi-Core Processors}}.
\newblock \bibinfo{journal}{\emph{IEEE Transactions on Computer-Aided Design of
  Integrated Circuits and Systems (TCAD)}} (\bibinfo{year}{2019}).
\newblock


\bibitem[\protect\citeauthoryear{{Wu} et~al\mbox{.}}{{Wu}
  et~al\mbox{.}}{2019}]%
        {facebook2019}
\bibfield{author}{\bibinfo{person}{C. {Wu}} {et~al\mbox{.}}}
  \bibinfo{year}{2019}\natexlab{}.
\newblock \showarticletitle{{Machine Learning at Facebook: Understanding
  Inference at the Edge}}. In \bibinfo{booktitle}{\emph{IEEE International
  Symposium on High Performance Computer Architecture (HPCA)}}.
\newblock


\bibitem[\protect\citeauthoryear{Xiao et~al\mbox{.}}{Xiao
  et~al\mbox{.}}{2018}]%
        {gandiva2018osdi}
\bibfield{author}{\bibinfo{person}{Wencong Xiao} {et~al\mbox{.}}}
  \bibinfo{year}{2018}\natexlab{}.
\newblock \showarticletitle{{Gandiva: Introspective Cluster Scheduling for Deep
  Learning}}. In \bibinfo{booktitle}{\emph{Proceedings of the 12th USENIX
  Conference on Operating Systems Design and Implementation (OSDI)}}.
\newblock


\bibitem[\protect\citeauthoryear{Xin, Tang, Lee, Yu, and Lin}{Xin
  et~al\mbox{.}}{2020}]%
        {deebert2020acl}
\bibfield{author}{\bibinfo{person}{Ji Xin}, \bibinfo{person}{Raphael Tang},
  \bibinfo{person}{Jaejun Lee}, \bibinfo{person}{Yaoliang Yu}, {and}
  \bibinfo{person}{Jimmy Lin}.} \bibinfo{year}{2020}\natexlab{}.
\newblock \showarticletitle{{{D}ee{BERT}: Dynamic Early Exiting for
  Accelerating {BERT} Inference}}. In \bibinfo{booktitle}{\emph{Proceedings of
  the 58th Annual Meeting of the Association for Computational Linguistics
  (ACL)}}. \bibinfo{publisher}{Association for Computational Linguistics},
  \bibinfo{pages}{2246--2251}.
\newblock


\bibitem[\protect\citeauthoryear{{Xing}, {Liang}, {Sui}, {Jia}, {Qiu}, {Liu},
  {Wang}, {Shan}, and {Wang}}{{Xing} et~al\mbox{.}}{2019}]%
        {dnnvm2019tcad}
\bibfield{author}{\bibinfo{person}{Y. {Xing}}, \bibinfo{person}{S. {Liang}},
  \bibinfo{person}{L. {Sui}}, \bibinfo{person}{X. {Jia}}, \bibinfo{person}{J.
  {Qiu}}, \bibinfo{person}{X. {Liu}}, \bibinfo{person}{Y. {Wang}},
  \bibinfo{person}{Y. {Shan}}, {and} \bibinfo{person}{Y. {Wang}}.}
  \bibinfo{year}{2019}\natexlab{}.
\newblock \showarticletitle{{DNNVM: End-to-End Compiler Leveraging
  Heterogeneous Optimizations on FPGA-based CNN Accelerators}}.
\newblock \bibinfo{journal}{\emph{IEEE Transactions on Computer-Aided Design of
  Integrated Circuits and Systems (TCAD)}} (\bibinfo{year}{2019}).
\newblock


\bibitem[\protect\citeauthoryear{Yang, Howard, Chen, Zhang, Go, Sandler, Sze,
  and Adam}{Yang et~al\mbox{.}}{2018}]%
        {netadapt2018eccv}
\bibfield{author}{\bibinfo{person}{Tien-Ju Yang}, \bibinfo{person}{Andrew
  Howard}, \bibinfo{person}{Bo Chen}, \bibinfo{person}{Xiao Zhang},
  \bibinfo{person}{Alec Go}, \bibinfo{person}{Mark Sandler},
  \bibinfo{person}{Vivienne Sze}, {and} \bibinfo{person}{Hartwig Adam}.}
  \bibinfo{year}{2018}\natexlab{}.
\newblock \showarticletitle{{NetAdapt: Platform-Aware Neural Network Adaptation
  for Mobile Applications}}. In \bibinfo{booktitle}{\emph{European Conference
  on Computer Vision (ECCV)}}.
\newblock


\bibitem[\protect\citeauthoryear{Zagoruyko and Komodakis}{Zagoruyko and
  Komodakis}{2017}]%
        {Zagoruyko2017}
\bibfield{author}{\bibinfo{person}{Sergey Zagoruyko} {and}
  \bibinfo{person}{Nikos Komodakis}.} \bibinfo{year}{2017}\natexlab{}.
\newblock \showarticletitle{{Paying more Attention to Attention: Improving the
  Performance of Convolutional Neural Networks via Attention Transfer}}. In
  \bibinfo{booktitle}{\emph{International Conference on Learning
  Representations (ICLR)}}.
\newblock


\bibitem[\protect\citeauthoryear{Zhang, Song, Gao, Chen, Bao, and Ma}{Zhang
  et~al\mbox{.}}{2019a}]%
        {Zhang_2019_ICCV}
\bibfield{author}{\bibinfo{person}{Linfeng Zhang}, \bibinfo{person}{Jiebo
  Song}, \bibinfo{person}{Anni Gao}, \bibinfo{person}{Jingwei Chen},
  \bibinfo{person}{Chenglong Bao}, {and} \bibinfo{person}{Kaisheng Ma}.}
  \bibinfo{year}{2019}\natexlab{a}.
\newblock \showarticletitle{{Be Your Own Teacher: Improve the Performance of
  Convolutional Neural Networks via Self Distillation}}. In
  \bibinfo{booktitle}{\emph{IEEE International Conference on Computer Vision
  (ICCV)}}.
\newblock


\bibitem[\protect\citeauthoryear{Zhang, Tan, Song, Chen, Bao, and Ma}{Zhang
  et~al\mbox{.}}{2019b}]%
        {scan2019neurips}
\bibfield{author}{\bibinfo{person}{Linfeng Zhang}, \bibinfo{person}{Zhanhong
  Tan}, \bibinfo{person}{Jiebo Song}, \bibinfo{person}{Jingwei Chen},
  \bibinfo{person}{Chenglong Bao}, {and} \bibinfo{person}{Kaisheng Ma}.}
  \bibinfo{year}{2019}\natexlab{b}.
\newblock \showarticletitle{{SCAN: A Scalable Neural Networks Framework Towards
  Compact and Efficient Models}}.
\newblock In \bibinfo{booktitle}{\emph{Advances in Neural Information
  Processing Systems (NeurIPS)}}.
\newblock


\end{thebibliography}
